\documentclass[11pt, letterpaper, logo, onecolumn, copyright]{main}

\usepackage[authoryear, sort&compress, round]{natbib}

\usepackage[inkscapeformat=png]{svg}

\usepackage[most, breakable, skins]{tcolorbox}

\tcbuselibrary{skins}
\usepackage{lipsum}
\usepackage{tabularx}
\usepackage{afterpage}
\usepackage{booktabs}
\usepackage{subcaption}
\usepackage{makecell}
\usepackage{multirow}
\usepackage{bm}
\usepackage{multicol}
\usepackage{array}
\usepackage{float}
\usepackage{listings, listings-rust}
\usepackage{fontawesome5}
\usepackage{hyperref}
\usepackage{amssymb,graphicx}
\usepackage[dvipsnames]{xcolor}
\usepackage{cleveref}
\usepackage{longtable}
\usepackage{pdflscape}
\usepackage{adjustbox}
\usepackage{nicematrix}
\usepackage{CJKutf8}
\usepackage{ragged2e}
\usepackage{colortbl}
\usepackage{enumitem}
\usepackage[ruled,linesnumbered]{algorithm2e}
\usepackage{pifont}
\usepackage[htt]{hyphenat}
\usepackage{amsmath}
\usepackage{amsthm}  
\usepackage{mathrsfs} 
\usepackage{circledsteps}
\usepackage{diagbox}
\usepackage{bookmark}

\usepackage{wrapfig}
\usepackage{xspace}
\usepackage{tikz}
\usepackage[normalem]{ulem}

\usepackage{pgfplots}
\usepackage{pgfplotstable}
\pgfplotsset{compat=1.18}

\definecolor{medgray55}{gray}{0.55}
\definecolor{medgray}{gray}{0.7}
\definecolor{litegray}{gray}{0.9}
\definecolor{gblue}{RGB}{210, 227, 252}
\definecolor{gred}{RGB}{250, 210, 207}
\definecolor{gyellow}{RGB}{254, 239, 195}
\definecolor{ggreen}{RGB}{206, 234, 214}
\definecolor{gorange}{RGB}{254, 223, 200}

\definecolor{gblue9}{RGB}{23, 78, 166}
\definecolor{gred9}{RGB}{165, 14, 14}
\definecolor{gyellow9}{RGB}{227, 116, 0}
\definecolor{ggreen9}{RGB}{13, 101, 45}
\definecolor{gorange9}{RGB}{176, 96, 0}

\definecolor{myblue}{rgb}{0,0,1}
\definecolor{myred}{rgb}{1,0,0}
\definecolor{mylightgray}{gray}{0.95}
\definecolor{myCite}{HTML}{1C4587}

\definecolor{highlightblue}{HTML}{185ABC}
\definecolor{cellHighlight}{HTML}{dbefff}

\usepackage{minitoc}

\noptcrule


\newcolumntype{L}[1]{>{\raggedright\let\newline\\\arraybackslash\hspace{0pt}}m{#1}}
\newcolumntype{C}[1]{>{\centering}m{#1}}

\newcolumntype{R}[1]{>{\raggedleft\let\newline\\\arraybackslash\hspace{0pt}}m{#1}}

\definecolor{ao}{rgb}{0.0, 0.0, 1.0}

\newcommand\vcent[1]{\vcenter{\hbox{#1}}}

\newcommand\loudspeaker[1][3]{\ensuremath{\vcent{\rule{.6ex}{.6ex}}\kern-.5ex
  \vcent{\scalebox{.6}[1]{\rotatebox[origin=center]{90}{$\blacktriangle$}}}
  \ifnum#1>0\relax\kern.05ex\vcent{\scalebox{.4}{\ttfamily)}}
  \ifnum#1>1\relax\kern-.4ex\vcent{\scalebox{.56}{\ttfamily)}}
  \ifnum#1>2\relax\kern-.55ex\vcent{\scalebox{.7}{\ttfamily)}}
  \fi\fi\fi}
}

\makeatletter
\renewcommand\subparagraph{
 \@startsection {subparagraph}{5}{\z@ }{3.25ex \@plus 1ex
 \@minus .2ex}{-1em}{\normalfont \normalsize \bfseries }}
\makeatother

\let\cite\citep
\hypersetup{
  citecolor = myCite,  
  linkcolor = myCite,   
  urlcolor  = myCite
}

\usepackage{amsmath}
\usepackage{amssymb}
\usepackage{mathtools}
\usepackage{amsthm}
\usepackage{graphicx}
\usepackage{amsmath} 
\usepackage{booktabs} 
\usepackage{enumitem}
\usepackage{times}  
\usepackage{courier}  
\usepackage{graphicx} 

\usepackage{algorithmic}
\usepackage{makecell}
\usepackage{booktabs}
\usepackage{multirow}
\usepackage{xcolor}
\usepackage{colortbl}
\usepackage{tcolorbox}
\usepackage{float}
\usepackage{amsmath, amssymb, amsfonts}

\title{DFPO: Scaling Value Modeling via Distributional Flow towards 
Robust and Generalizable LLM Post-Training}
\author{
    Dingwei Zhu$^1$,  Zhiheng Xi$^1$, Shihan Dou$^1$, Jiahan Li$^1$, Chenhao Huang$^1$,  \\
\textbf{Junjie Ye$^1$, Sixian Li$^1$,  Mingxu Chai$^1$, Yuhui Wang$^1$, Yajie Yang$^1$,Ming Zhang$^1$} \\
\textbf{ Jiazheng Zhang$^1$, Shichun Liu$^1$, Caishuang Huang$^1$, Yunke Zhang$^2$,} \\
\textbf{  Yuran Wang$^2$,Tao Gui$^1$$^\dag$, Xipeng Qiu$^1$, Qi Zhang$^1$, Xuanjing Huang$^1$}
\\
$^1$College of Computer Science and Artificial Intelligence, Fudan University \\$^2$Honor Device Co., Ltd \\
\texttt{dwzhu25@m.fudan.edu.cn, tgui@fudan.edu.cn} 
}
\vspace{-50pt}

\begin{abstract}
Training reinforcement learning (RL) systems in real-world environments remains challenging due to noisy supervision and poor out-of-domain (OOD) generalization, especially in LLM post-training. Recent distributional RL methods improve robustness by modeling values with multiple quantile points, but they still learn each quantile independently as a scalar. This results in rough-grained value representations that lack fine-grained conditioning on state information, struggling under complex and OOD conditions. We propose \textbf{DFPO} (Distributional Value Flow Policy Optimization with Conditional Risk and Consistency Control), a robust distributional RL framework that models values as continuous flows across time steps. By scaling value modeling through learning of a value flow field instead of isolated quantile predictions, DFPO captures richer state information for more accurate advantage estimation. To stabilize training under noisy feedback, DFPO further integrates conditional risk control and consistency constraints along value flow trajectories. Experiments on dialogue, math reasoning, and scientific tasks show that DFPO outperforms PPO, FlowRL, and other robust baselines under noisy supervision, achieving improved training stability and generalization.
\end{abstract}

\begin{document}

\doparttoc
\faketableofcontents

\begingroup
  \renewcommand\thefootnote{}
  \footnote{
            \textsuperscript{\dag}Corresponding authors.}
  \addtocounter{footnote}{-1}
\endgroup

\vspace{-30pt}
\maketitle

\begin{figure}[!ht]
\begin{center}
\vspace{0pt}
\includegraphics[width=1\linewidth]{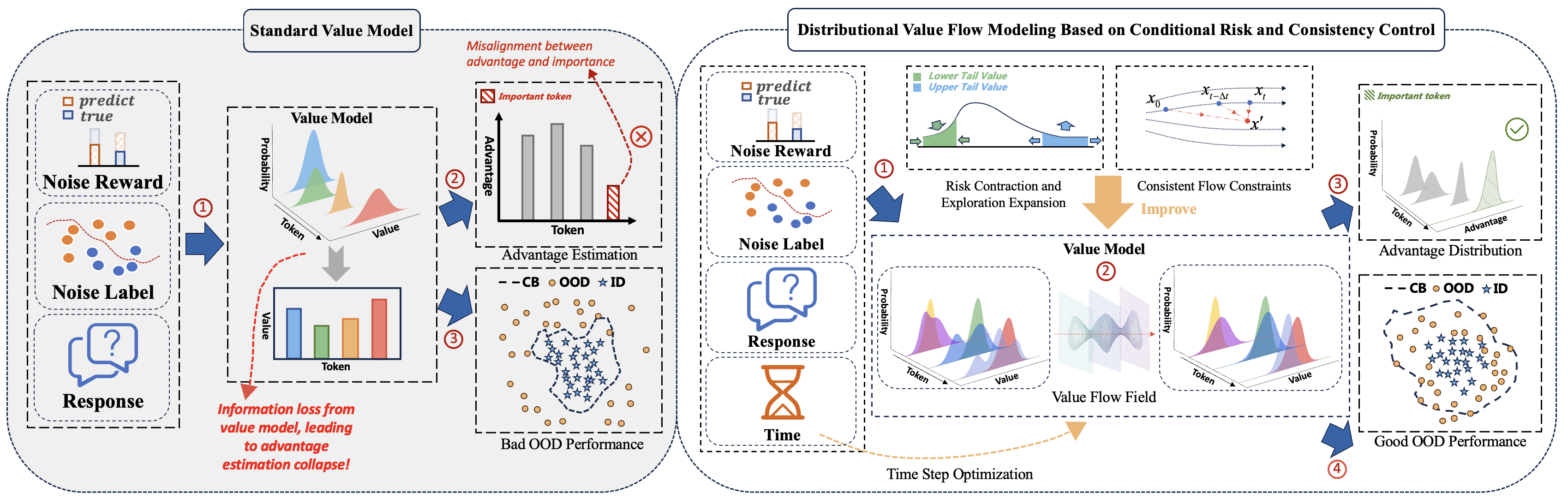}
\end{center}
\vspace{-16pt}
\caption{Comparison between the standard scalar value model and DFPO with distributional value flow modeling.
The standard value model is sensitive to noisy and biased reward signals, which often leads to unstable value estimation and unreliable advantage learning.
In contrast, DFPO models a token-level distributional value flow field across time steps, thereby capturing state information more effectively under noisy supervision. By integrating conditional risk control and consistency constraints along value flow trajectories, DFPO suppresses spurious value fluctuations while preserving high-value exploration. CB denotes the capacity boundary of the model.}
\label{fig:all}

\end{figure}

\section{Introduction}

Reinforcement Learning has achieved remarkable success across diverse application domains \cite{xi2025agentgymrltrainingllmagents,xi2024agentgymevolvinglargelanguage,xi2024traininglargelanguagemodels,ding2025mitigatingtailnarrowingllm,2024arXiv240219128C}. However, real-world RL training remains challenging due to severe instability and poor cross-domain generalization, primarily caused by pervasive noise and incomplete supervision \cite{geng2024noisedistributiondecompositionbased,Para__2024,zhu2025vrporethinkingvaluemodeling}. This issue is particularly pronounced in LLM post-training and agent-based RL, where learning relies on noisy, high-variance, semantically complex feedback for human value alignment or self-improvement.

To address these challenges, prior work has explored advantage control-based RL methods \cite{wang2025lambdagrpounifyinggrpoframeworks}. Clipping-based approaches \cite{xi2025bapostabilizingoffpolicyreinforcement,liu2025cpgdstablerulebasedreinforcement} stabilize policy updates by reshaping or truncating noisy advantage estimates, while methods like KTAE \cite{sun2025ktaemodelfreealgorithmkeytokens} reweight signals via semantic key words to suppress spurious noise. Despite effective variance reduction, these methods focus on stabilizing advantage fluctuations rather than explicitly guiding OOD learning, limiting their generalization in real-world OOD scenarios.

Recent studies adopt distributional value models \cite{zhu2025dvpodistributionalvaluemodelingbased} and some distributionally robust Bellman methods \cite{ma2021conservativeofflinedistributionalreinforcement,hu2025value} to enhance robustness under noise and OOD conditions. By modeling value supervision across multiple dimensions, these methods capture more comprehensive state and uncertainty information, improving advantage estimation in complex environments. However, they typically rely on discrete scalar quantile learning for each value dimension, leading to rough-grained state information conditioning and impeding the generalization of value predictions in complex, dynamic OOD scenarios.

In this work, we introduce flow-based modeling into robust distributional RL. Instead of relying on isolated quantile points, we further scale value modeling by representing the value distribution for each state as a continuous generative flow process defined over a virtual time horizon. Learning value distributions as a continuous flow field enables more expressive semantic representations and finer-grained uncertainty modeling. We impose principled constraints inspired by conditional risk and consistency theories on this generative flow field, which ensures optimal transport under noisy and OOD conditions and enables accurate advantage estimation.

We propose \textbf{DFPO} (Distributional Value Flow Policy Optimization with Conditional Risk and Consistency Control), a robust RL framework integrating continuous value flow modeling with risk and consistency control. DFPO further scales value modeling by representing token-level values as continuous distributions via flow dynamics. This approach encourages high-value exploration while suppressing lower-tail risk, and enforces consistency constraints along the virtual flow paths to stabilize the generative process and guide coherent predictions.

Evaluations on multi-turn dialogue, math reasoning, and scientific QA tasks show that DFPO consistently outperforms standard PPO, GRPO, distributionally robust Bellman methods, and FlowRL under noisy and OOD settings. By combining continuous flow supervision with risk and consistency control, DFPO achieves superior training stability and generalization, providing a scalable solution for real-world robust RL.
Our main contributions are summarized as follows:
\begin{itemize}
    \item We propose a distributional value flow framework that scales value modeling by replacing discrete quantile learning with continuous generative flow modeling over a virtual horizon. 
    \item We introduce a conditional risk and consistency flow control mechanism, jointly optimizing bounded risk distributions and trajectory coherence to balance exploration and constraint and improve training stability.
    \item Extensive experiments demonstrate DFPO’s superiority under noisy and OOD conditions, offering a scalable solution for real-world robust RL.
\end{itemize}

\begin{figure*}[h]
\centering
\includegraphics[width=1\linewidth]{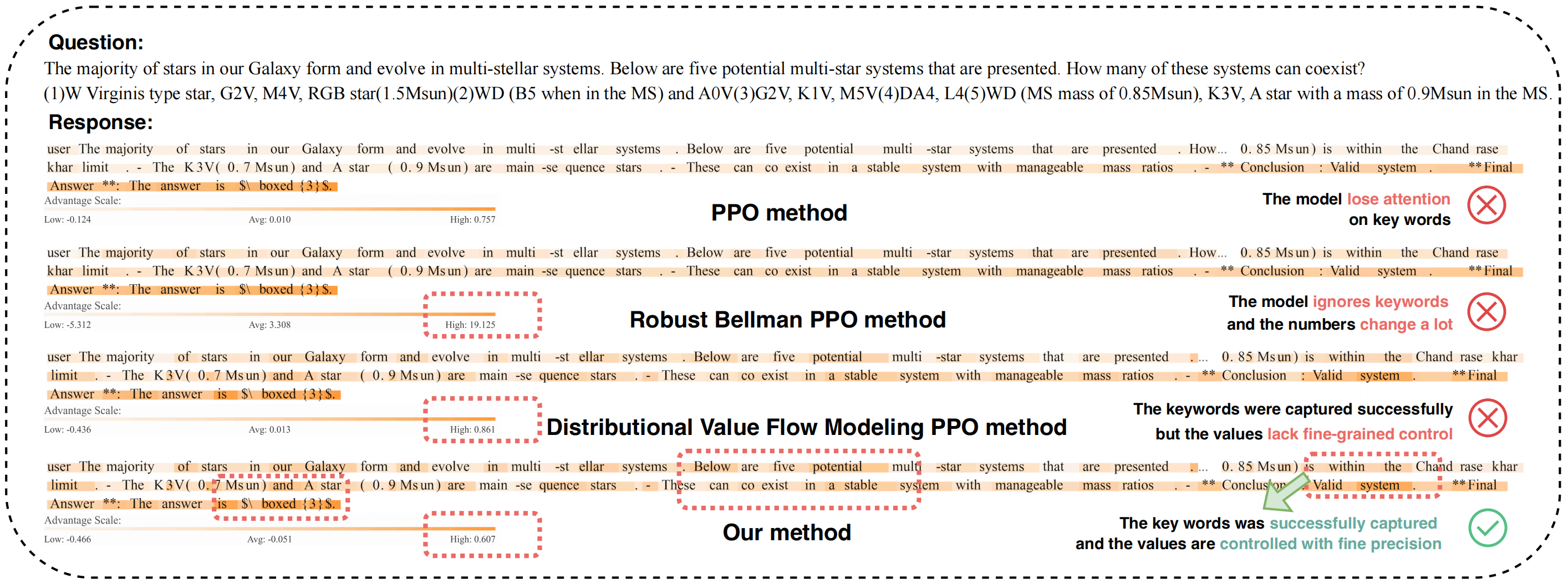}
\caption{Token-level advantage estimation for the same response across different methods. Our method demonstrates better alignment between high advantage scores and key words.}
\label{fig:adv-vis}
\end{figure*}

\section{Related Work}

\paragraph{Distributional Reinforcement Learning}
Distributional Reinforcement Learning (DRL) extends traditional reinforcement learning by modeling the full probability distribution of future returns rather than just their expectation. It provides a richer signal for capturing higher-order statistics, including variance, skewness, and multimodality, to enhance decision stability and robustness. Early works like C51, QR-DQN, and IQN~\cite{bellemare2017distributionalperspectivereinforcementlearning,dabney2017distributionalreinforcementlearningquantile,dabney2018implicitquantilenetworksdistributional} attempted to model such full distributions.
Recent RL from Human Feedback (RLHF) advances adopt distributional ideas.  Q\# \cite{zhou2025qsharpprovablyoptimaldistributional} applies DRL to LLM post-training, computing a provably optimal KL-regularized Q-function via the reference policy’s cumulative return distribution.
Current distribution constraint methods use distribution distances and robust Bellman-based pessimistic optimization. Kan et al. \cite{sun2024distributionalreinforcementlearningregularized} controls multi-distribution divergence via Sinkhorn Divergence and gradients; VDRL \cite{hu2025value} mitigates Q-value overestimation by discarding lowest multi-Q estimates; CODAC \cite{ma2021conservativeofflinedistributionalreinforcement} penalizes quantiles to learn point-wise lower bounds for conservatism.
Building on these, this work introduces distributional supervision and conditional value risk constraints into value modeling to improve stability and generalization under noisy supervision.

\paragraph{Flow Reinforcement Learning}

Flow Reinforcement Learning extends traditional continuous control and value estimation by leveraging flow-based generative models to model complex, multimodal distributions beyond Gaussian approximations. It offers superior expressivity for capturing high-dimensional action policies and return distributions while mitigating the sampling overhead of diffusion models \cite{ghugare2025normalizingflowscapablemodels}.
In policy optimization, FPO \cite{mcallister2025flowmatchingpolicygradients}, FlowRL \cite{zhu2025flowrlmatchingrewarddistributions} integrate CFM into gradient frameworks for multi-modality; ReinFlow \cite{zhang2026reinflowfinetuningflowmatching} and Riemannian Flow Matching \cite{braun2024riemannianflowmatchingpolicy} add noise injection and geometric constraints for robotic task exploration and consistency.
For value estimation, recent works focus on distributional fidelity and scalability. Value Flows \cite{dong2025valueflows} models return distributions via density paths and flow derivative ODEs for uncertainty quantification; floq \cite{agrawalla2025floqtrainingcriticsflowmatching} parameterizes Q-functions as velocity fields for test-time scaling.
To ensure robustness in offline and embodied settings, Akimov et al. \cite{akimov2023letofflinerlflow} enforces latent-space conservatism; FM-IRL \cite{wan2025fmirlflowmatchingrewardmodeling} unifies reward modeling and policy regularization.
Building on these, DFPO integrates flow learning into value distribution modeling, enabling fine-grained generalization and robustness control via state guidance.

\begin{figure*}[h]
\centering
\includegraphics[width=1\linewidth]{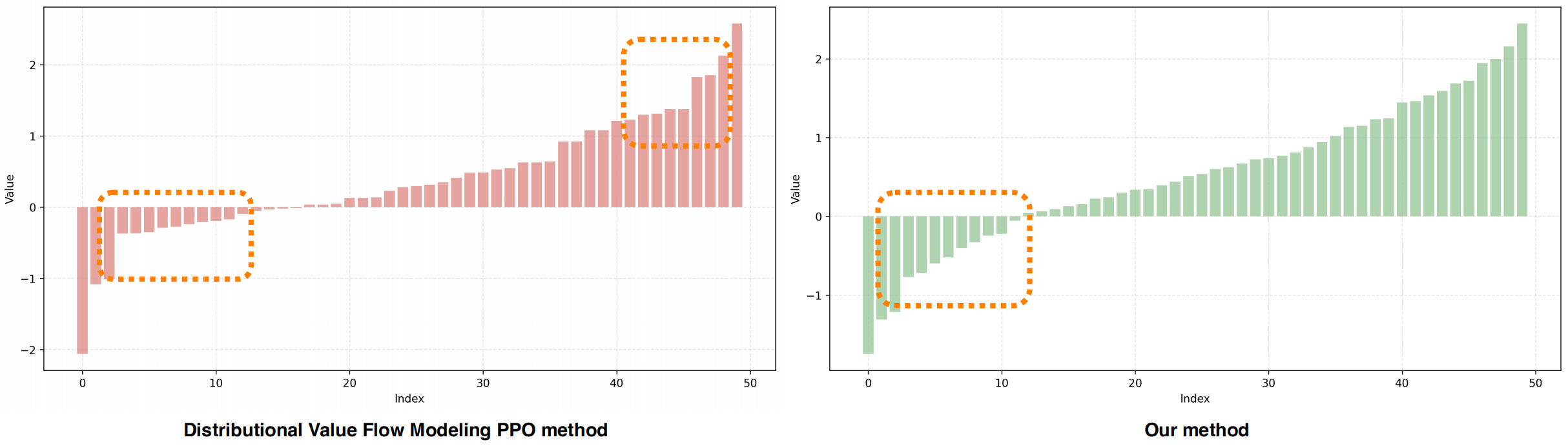}
\caption{\textbf{Comparison of the output value distributions of one token for the answer part.}
(Left) The standard distributional value flow modeling PPO method shows sharp, unstable lower-tail expansion, indicating excessive risk accumulation and unreliable training variance.
(Right) DFPO constrains lower-tail risk (slight contraction) while promoting upper-tail exploratory expansion, balancing noise robustness and generalization in complex scenarios.}
\label{fig:adv-fenbu}
\end{figure*}

\begin{figure*}[h]
\centering
\includegraphics[width=1\linewidth]{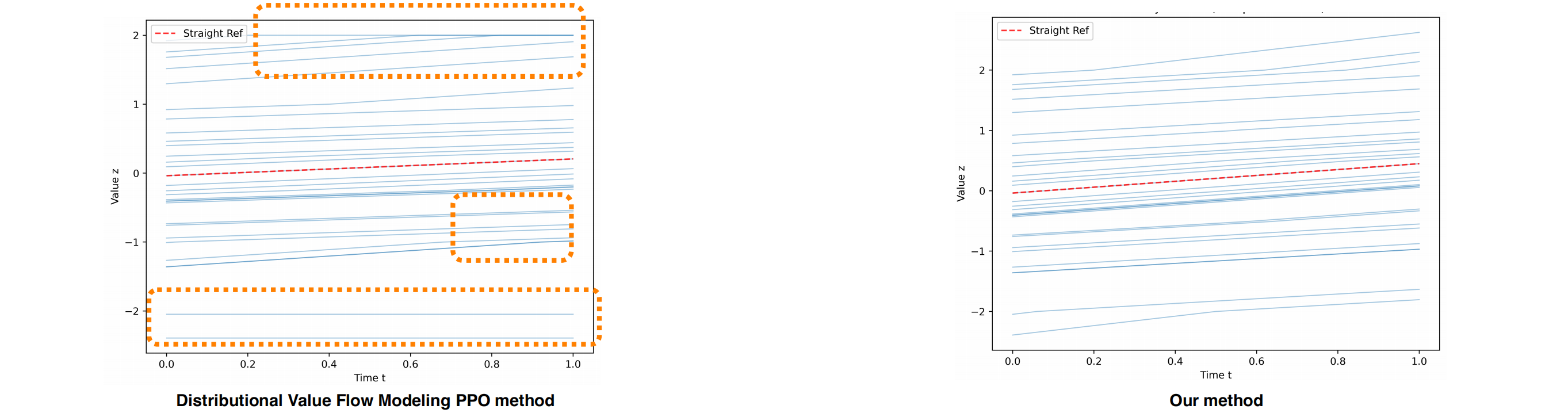}
\caption{\textbf{Comparison of value flow trajectories of one token for the answer part.}
(Left) The standard distributional value flow modeling PPO method without consistency constraints generates intertwined, chaotic trajectories, causing unstable updates and poor generalization.
(Right) DFPO produces smooth, coherent paths for accurate advantage estimation even under OOD conditions.}
\label{fig:adv-flow}
\end{figure*}

\section{Method}

\subsection{Motivation}
To enhance the generalization and robustness of RL value models in complex real-world scenarios, prior work~\cite{ma2021conservativeofflinedistributionalreinforcement,hu2025value} models values via dense distributions with multiple quantile points. These methods aim for more complete state descriptions and accurate advantage estimation under noisy supervision or OOD settings. However, existing distributional methods typically learn each quantile independently as a scalar. This results in rough-grained representations that hinder the generalization of value predictions in dynamic environments, ultimately limiting performance under OOD conditions.
To address this limitation, we further scale value modeling to guide the model's training in real environments. Specifically, we extend value learning beyond discrete scalar quantiles by introducing flow-based learning, which models continuous value flows across time steps for each state. This expressive flow field explicitly controls value flow divergence and convergence, boosting generalization in OOD scenarios by capturing rich temporal dynamics and semantic structures inaccessible to discretized quantile methods.

To further scale value modeling and model the value flow field across time, standard flow matching methods~\cite{liu2022flowstraightfastlearning} typically sample virtual time steps at random during training to enable learning over entire trajectories. However, this random sampling introduces significant noise into the learned flows~\cite{zhang2026reinflowfinetuningflowmatching,braun2024riemannianflowmatchingpolicy}. As a result, flow directions often become unstable, leading to large variations in the generative paths and major instability in the resulting value distributions.
As shown in Figure~\ref{fig:adv-flow}, a PPO-based baseline applying value distribution modeling and flow learning without additional constraints produces heavily overlapping and intertwined trajectories during the generative process. Furthermore, as shown in Figure~\ref{fig:adv-fenbu}, the value distributions exhibit sharp and unstable increases in the lower tail, reflecting excessive risk accumulation and unreliable advantage estimates.

Inspired by consistency-based learning~\cite{lu2025simplifyingstabilizingscalingcontinuoustime}, we mitigate these issues by anchoring the current flow to both past flow information and the current state. By enforcing directional consistency at multiple points along each virtual flow trajectory, we constrain the generative evolution toward more stable and coherent paths. We simultaneously incorporate explicit risk control in distribution modeling, mitigating risk fluctuations while preserving high-value exploration.
As illustrated in Figure~\ref{fig:adv-flow}, DFPO produces smoother flow trajectories. Consequently, it achieves a more stable distributional evolution, as shown in Figure~\ref{fig:adv-fenbu}, effectively contracting lower-tail risk while expanding upper-tail exploration. This leads to more reliable advantage estimation under noisy and OOD conditions, enabling the model to consistently capture key words. Overall, DFPO achieves a superior balance between exploration and constraint, effectively extending both generalization and robustness in complex environments.

\subsection{Auto-regressive Flow-based Value Modeling}
\label{sec:architecture}

To capture the complex, multi-modal distribution of future returns, we construct a distributional value model $V_\theta$ on top of a pre-trained auto-regressive transformer backbone. Let $\mathbf{s} = (w_1, \dots, w_L)$ denote an input sequence of length $L$, and $\mathbf{h}_{\mathbf{s}} \in \mathbb{R}^d$ denote the corresponding hidden state representation generated by the transformer.

Unlike traditional approaches that predict a scalar value or a static categorical distribution, we introduce a Value Flow Head, parameterized as a continuous-time neural Ordinary Differential Equation (ODE) vector field $v_\theta$. This vector field defines the evolution of the value distribution's path over a virtual time horizon $t \in [0, 1]$.

The architecture processes inputs sequentially. First, the time step $t$ is mapped to a high-dimensional embedding using sinusoidal positional encodings, followed by a Multi-Layer Perceptron (MLP) with Mish activations, denoted as $\phi(t)$. This embedding is then conditioned on the hidden state $h_s$ and the noisy value input $z$ via concatenation. Finally, to ensure the Lipschitz continuity of the learned vector field and stabilize the flow dynamics, we apply spectral normalization to the linear layers within the value head.
Formally, the output of the value head approximates the velocity of the probability path: $v_\theta(z_t, t, \mathbf{h}_{\mathbf{s}}) \approx \frac{d z_t}{dt}$

During inference, the return distribution is approximated by sampling an initial noise batch $\{z_0^{(k)}\}_{k=1}^K \sim \mathcal{N}(0, 1)$ and solving the initial value problem (IVP) from $t=0$ to $t=1$ using an Euler solver~\cite{liu2022flowstraightfastlearning}. The terminal particles $\{z_1^{(k)}\}_{k=1}^K$ represent the empirical quantiles of the predicted return distribution.


\subsection{Distributional Generalized Advantage Estimation}
\label{sec:d_gae}

To effectively train the distributional critic, we extend the Generalized Advantage Estimation (GAE) to the distributional setting. Traditional GAE operates on scalar expectations, which loses information regarding the aleatoric uncertainty of the environment. We introduce a Distributional GAE Processor that operates directly in the quantile space.

Let $Z(s)$ denote the random variable representing the return at state $s$. We represent distributions using a fixed set of quantile positions $\tau = \{\tau_1, \dots, \tau_K\}$. The distributional temporal difference (TD) error at time step $t$, denoted as $\delta_{D, t}$, is defined as a distribution $\delta_{D, t} := R_t + \gamma Z(s_{t+1}) - Z(s_t)$, where $R_t$ is the scalar reward (broadcasted to the distribution dimension), $\gamma \in [0, 1]$ is the discount factor, and arithmetic operations are performed element-wise on the sorted quantile supports.

The Distributional Advantage $A_D(s_t)$ is calculated recursively as $A_D(s_t) = \delta_{D, t} + (\gamma \lambda) A_D(s_{t+1})$. Here, $\lambda \in [0, 1]$ serves as the smoothing parameter for the distributional variance. The scalar multiplication by $\gamma \lambda$ scales the spread of the distribution around its mean. This mechanism propagates the full distributional shape of value estimates through the trajectory, providing fine-grained supervision for risk-sensitive optimization.

\subsection{Robust Training via Risk-Sensitive Flow Matching}
\label{sec:training}

We train the value model using a composite objective function that combines Uncertainty-Weighted Distributional Conditional Flow Matching (UDCFM) for generative modeling with geometric and risk-based constraints for robustness.

\subsubsection{Robust Distributional Flow Learning}

To effectively model complex return distributions in high-dimensional environments, we propose a robust training paradigm that anchors the flow field to stable estimates and enforces geometric consistency.

\paragraph{Uncertainty-Weighted Distributional CFM (UDCFM)}
The primary objective regresses the neural vector field $v_\theta$ to a target probability path.  Given a source distribution $p_0 = \mathcal{N}(0, 1)$ and a target return sample $x_1$ derived from the GAE buffer, we define the optimal transport path as $z_t = t \cdot x_1 + (1-t) \cdot x_0$.  The target velocity is $u_t(z_t | x_1, x_0) = x_1 - x_0$.
To mitigate the impact of noisy supervision, we introduce a state-dependent confidence weight $w_{\text{conf}}(\mathbf{s})$.  We approximate the flow sensitivity by estimating the Jacobian evolution $\|J(1)\|^2$ via the sensitivity ODE.  The weight is derived as $w_{\text{conf}}(\mathbf{s}) = \sigma(\|J(1)\|^2 / \tau_{\text{temp}}) + 0.5$, where $\sigma$ is the sigmoid function and $\tau_{\text{temp}}$ is a temperature scaling factor.  The weighted UDCFM objective is:
\begin{equation}\small
\mathcal{L}_{\text{UDCFM}}(\theta) = \mathbb{E}_{t \sim \mathcal{U}[0,1], x_0, x_1, \mathbf{s}} [ w_{\text{conf}}(\mathbf{s}) \cdot
\left\| v_\theta(z_t, t, \mathbf{h}_{\mathbf{s}}) - (x_1 - x_0) \right\|^2 ]
\end{equation}

\paragraph{Bootstrapped Anchor Regularization (BCFM)}
In the early stages of training, the high variance of the stochastic target $x_1$ can destabilize the learning of the vector field.  To mitigate this, we implement a Bootstrapped CFM term using an "anchor" target $x_{1}^{\text{anc}} = \text{StopGrad}(\mathbb{E}_{\theta}[x_1 | \mathbf{s}])$, which represents the current stable estimate of the expected return.  Here, the operator $\text{StopGrad}(\cdot)$ denotes the stop-gradient operation (or detach), which blocks gradient backpropagation through the target term.  This ensures that $x_{1}^{\text{anc}}$ serves as a fixed regression target during the update step, preventing the optimization from collapsing into trivial solutions where the target shifts to match the prediction.  A secondary flow loss minimizes the deviation towards this anchor:
\begin{equation}\small
\mathcal{L}_{\text{BCFM}}(\theta) = \mathbb{E}_{t \sim \mathcal{U}[0,1], x_0, \mathbf{s}}[ w_{\text{conf}}(\mathbf{s}) \cdot \\ \left\| v_\theta(z_t^{\text{anc}}, t, \mathbf{h}_{\mathbf{s}}) - (x_{1}^{\text{anc}} - x_0) \right\|^2 ]
\end{equation}
where $z_t^{\text{anc}}$ is the interpolated state along the anchor path.

\paragraph{Geometric Consistency Regularization}
To ensure the learned vector field constitutes a mathematically consistent ODE trajectory, we enforce a geometric consistency constraint. By minimizing the consistency loss, we implicitly encourage the flow trajectories to approximate Optimal Transport paths. This geometric rectification is crucial for enabling accurate single-step inference during the policy optimization phase. Under the linear flow assumption, the terminal state $x_1$ can be projected from any intermediate state $z_t$ via $\hat{x}_1(z_t, v_t) = z_t + (1-t) \cdot v_t$. We employ Symmetric Time Sampling: for $t \sim \mathcal{U}[0, 1]$ and its conjugate $t' = 1 - t$, we enforce that projections yield an identical terminal value:
\begin{equation}\small
    \mathcal{L}_{\text{cons}}(\theta) = \left\| \hat{x}_1(z_t, v_\theta(z_t)) - \hat{x}_1(z_{t'}, v_\theta(z_{t'})) \right\|^2
\end{equation}

\subsubsection{Risk-Controlled Distributional Constraints}

To enhance policy robustness against OOD states, we impose explicit constraints on the shape of the learned distribution. Let $\{\hat{z}_{(k)}\}_{k=1}^K$ be the sorted predicted quantiles and $\{z^{\text{tgt}}_{(k)}\}_{k=1}^K$ be the sorted target quantiles (target returns).

\paragraph{Conditional Value Risk  Optimization}
We enforce a risk-averse constraint on the left tail and a gain constraint on the right tail. For a risk level $\alpha \in (0, 1)$ and gain level $\beta \in (0, 1)$, we define the index cutoffs $k_\alpha = \lfloor \alpha K \rfloor$ and $k_\beta = \lfloor (1-\beta) K \rfloor$. The risk loss $\mathcal{L}_{\text{Risk}}$ is defined as:
\begin{equation}
\small
\mathcal{L}_{\text{Risk}} = \left\| \frac{1}{k_\alpha} \sum_{k=1}^{k_\alpha} (\hat{z}_{(k)} - z^{\text{tgt}}_{(k)}) \right\|^2 + \bigg\| \frac{1}{K -k_\beta+1} \sum_{k=k_\beta}^{K} (\hat{z}_{(k)} - z^{\text{tgt}}_{(k)}) \bigg\|^2
\end{equation}

\paragraph{Tail Curvature and Shape Regularization}
To balance robustness and generalization, we penalize the second-order finite differences $\nabla^2 \hat{z}_{(k)} = \hat{z}_{(k+2)} - 2\hat{z}_{(k+1)} + \hat{z}_{(k)}$ in the tail regions $\mathcal{I}_{L}$ (left) and $\mathcal{I}_{R}$ (right). We define the tail index sets as $\mathcal{I}_{L} = \{1, \dots, k_\alpha - 2\}$ for the left tail and $\mathcal{I}_{R} = \{k_\beta, \dots, K - 2\}$ for the right tail. We enforce concavity in the left tail and convexity in the right tail using the ReLU operator $[\cdot]_+$:
\begin{equation}\small
    \mathcal{L}_{\text{Shape}} = \frac{1}{|\mathcal{I}_L|} \sum_{k \in \mathcal{I}_L} \left[ \nabla^2 \hat{z}_{(k)} \right]_+ + \frac{1}{|\mathcal{I}_R|} \sum_{k \in \mathcal{I}_R} \left[ -\nabla^2 \hat{z}_{(k)} \right]_+
\end{equation}

\subsubsection{Total Objective}
The final objective function for the value model combines the generative flow modeling with distributional robustness constraints:
\begin{equation}\small
    \mathcal{L}_{\text{Total}} = \mathcal{L}_{\text{UDCFM}} + \lambda_{\text{reg}}\mathcal{L}_{\text{BCFM}} + \lambda_{\text{cons}}\mathcal{L}_{\text{cons}} + 
   \lambda_{\text{risk}}\mathcal{L}_{\text{Risk}} + \lambda_{\text{shape}}\mathcal{L}_{\text{Shape}}
\end{equation}
where $\lambda_{\text{reg}}, \lambda_{\text{cons}}, \lambda_{\text{risk}}, \lambda_{\text{shape}}$ are hyperparameters balancing the regularization terms.

\section{Experiments}
\subsection{Experimental Setup}

We evaluate DFPO on dialogue, mathematical reasoning, and scientific QA tasks under both model-based and rule-based reward supervision.
All experiments follow a unified training pipeline to ensure fair comparison across methods, while the learning objectives and value modeling differ across approaches.

\paragraph{Dialogue Task}
To evaluate DFPO under realistic model-based reward supervision, we conduct multi-turn dialogue experiments on the \textbf{Honor-Dialogue Dataset}.
This dataset consists of task-oriented, multi-domain conversations collected from real-world usage scenarios.
The model is trained to act as a dialogue assistant that produces natural, helpful, and task-completing responses.
Reinforcement learning is performed using noisy feedback from a dialogue reward model on real multi-turn trajectories.
Evaluation is carried out through interactions with GPT-4o, following the dialogue evaluation methods described in Section~4.2.
All data have been anonymized, and the dataset usage has been reviewed and approved by the ethics committee.

\paragraph{Math and Science Tasks}
To assess robustness under rule-based supervision, we further evaluate DFPO on mathematical and scientific reasoning tasks.
We adopt the Light-R1 dataset~\cite{wen2025lightr1curriculumsftdpo} for math training and SuperGPQA~\cite{pteam2025supergpqascalingllmevaluation} for scientific training.
For each dataset, we first generate multiple candidate solutions using the initial model and apply majority voting to obtain pseudo-labels for reinforcement learning.
Evaluation is conducted on a diverse set of benchmarks, including math datasets
MATH500~\cite{hendrycks2021measuringmathematicalproblemsolving}, AIME24, Minerva-Math~\cite{lewkowycz2022solvingquantitativereasoningproblems}, AMC23
and science datasets
SampleQA~\cite{wei2024measuringshortformfactualitylarge}, GPQA, and HLE.
Due to memory constraints, generation is performed over multiple rounds, with a maximum length of 4096 tokens per round, and the final round output is used for evaluation.

\paragraph{Model Initialization and Baselines}
For math and science tasks, DFPO and all baselines are initialized from Qwen3-8B~\cite{qwen3}.
For dialogue tasks, both the policy model and reward model are initialized from Qwen3-8B and further fine-tuned on the Honor-Dialogue dataset.
In addition to robust RL baselines like KTAE, BAPO, and $\lambda$-GRPO ~\cite{sun2025ktaemodelfreealgorithmkeytokens,wang2025lambdagrpounifyinggrpoframeworks,xi2025bapostabilizingoffpolicyreinforcement}, we include flow-based RL methods like FlowRL~\cite{zhu2025flowrlmatchingrewarddistributions} and robust distributional value RL methods that construct multiple value estimates per token and aggregate them to guide policy optimization under noisy rewards. Since no relevant work exists in the LLM domain, we implemented a distributional PPO algorithm based on the robust Bellman operator inspired by recent studies~\cite{ma2021conservativeofflinedistributionalreinforcement,hu2025value}. This algorithm simulates a worst-case learning scenario by selecting the minimum value from the distributional predictions over 200 dimensions. 
\begin{figure}[h]
\centering
\includegraphics[width=0.6 \linewidth]{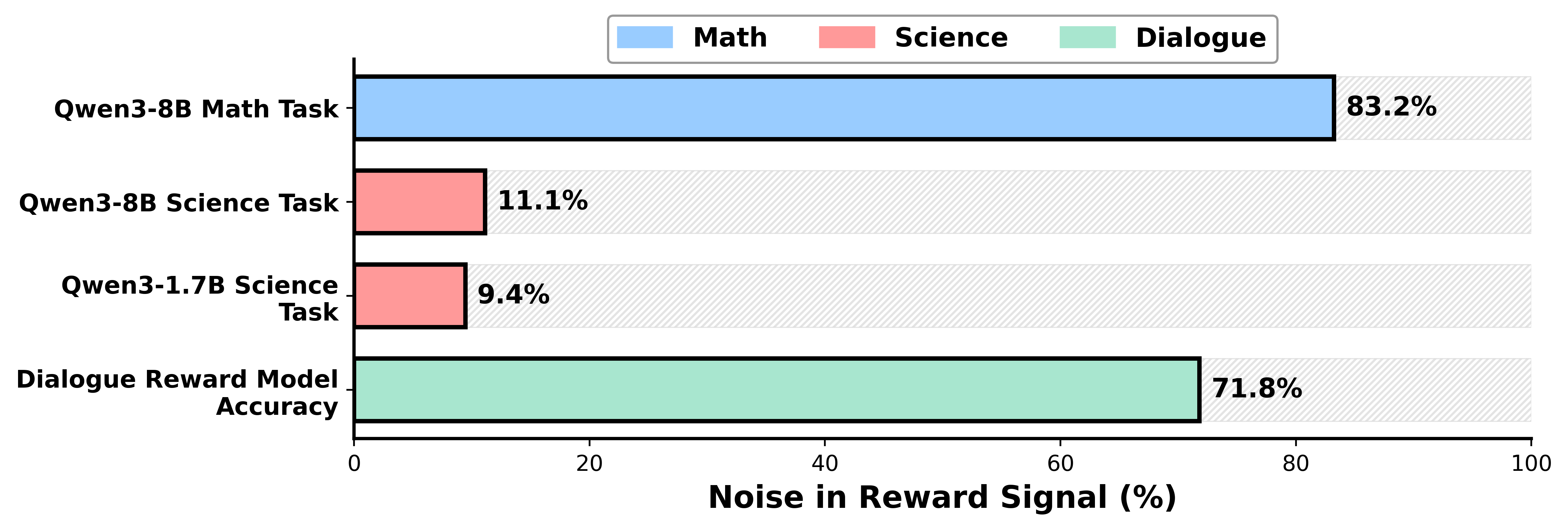}
\caption{Noise statistics in the various tasks. A significant portion of rewards contains inaccuracies.}
\label{fig:task-noise}
\end{figure}

\paragraph{Training Noise}
Noise is pervasive in both model-based and rule-based reward supervision. Figure~\ref{fig:task-noise} shows the error rates of different reward signals. Specifically, rule-based rewards from majority-voted generations contain mislabels, causing erroneous supervision under noisy scenarios in math and science tasks.

\paragraph{Implementation Details}
Unless otherwise specified, majority voting is performed over 5 samples.
During the exploration phase, we employ a sampling temperature of 1.0 with a top-$p$ setting of 1.0. For the optimization objective, the hyperparameters balancing the loss components are configured as follows: the regularization weights are set to $\lambda_{\text{reg}}=0.1$ for Bootstrapped CFM and $\lambda_{\text{cons}}=0.01$ for geometric consistency; the distributional constraint weights are set to $\lambda_{\text{risk}}=0.5$, and $\lambda_{\text{Shape}}=0.5$.
In the distributional value model, the risk level $\alpha$ and gain level $\beta$ are both set to 0.1 (corresponding to the $10\%$ and $90\%$ quantiles). The number of flow sampling steps (quantiles) is set to $K=50$, and the Jacobian estimation for confidence weighting utilizes 10 ODE steps. The model generates sequences of length 4096 for reasoning tasks and 1024 for dialogue tasks. The reinforcement learning training is performed for 1 epoch per iteration on a cluster of 8 $\times$ NVIDIA A100 80GB GPUs.

\begin{figure}[h]
    \centering
    \includegraphics[width=0.85 \linewidth]{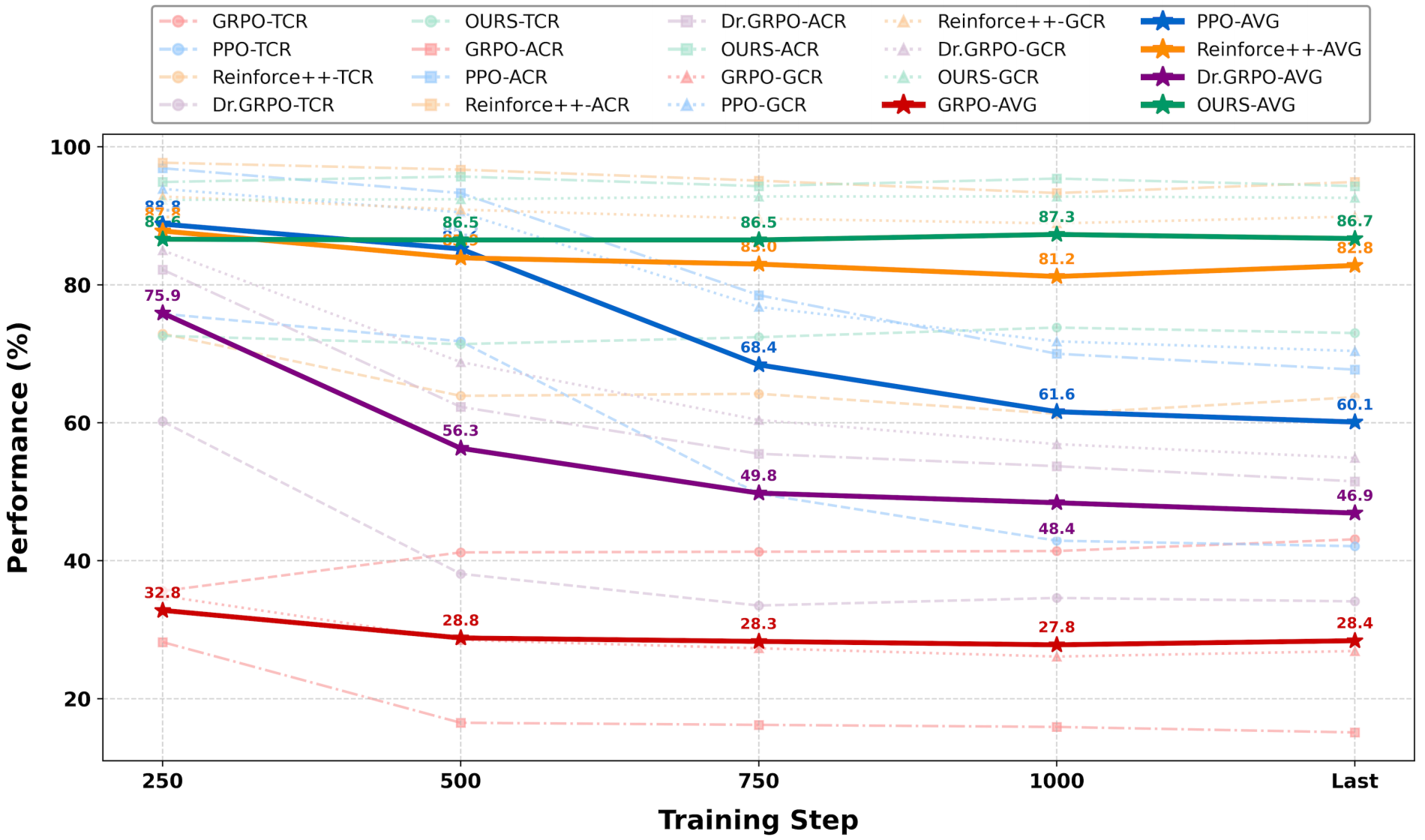}
    \caption{Training performance on dialogue tasks. DFPO demonstrates exceptional stability, maintaining a high average performance of 86.7\% throughout the training process. }
\label{fig:dialogue_traj}
\end{figure}
\subsection{Evaluation Metrics}
For mathematical reasoning and scientific QA tasks, we report accuracy as the primary evaluation metric.
For multi-turn dialogue tasks, we adopt a three-part evaluation methods:
(1) Task Completion Rate (TCR), measuring whether the intended task is successfully completed;
(2) Ask Completion Rate (ACR), evaluating whether key user requests are fully addressed; 
(3) Goal Completion Rate (GCR), assessing like overall fluency, coherence, logicality and response quality and so on.
This evaluation framework allows us to measure both task effectiveness and alignment with human communication standards. Detailed evaluation methods are provided in the Appendix.

\subsection{Experiment results}
\begin{table*}[t]
\centering
\small
\caption{
Performance comparison (\%) of different reinforcement learning methods on \textbf{real-world dialog domains} at 1000 training steps.
DFPO achieves the highest overall average accuracy 87.31\% and maintains robust generalization across domains, effectively \textbf{preventing the catastrophic forgetting} observed in other baselines.
In contrast, GRPO and Dr.GRPO suffer from severe performance collapse ($<$50\%) at the end of training. PPO also exhibits significant instability, dropping to $\sim$60\% average accuracy.
Notably, under the influence of real-world noise, the Robust Bellman PPO, BAPO and FlowRL methods generated outputs with \textbf{formatting errors}, making performance evaluation infeasible.
}
\resizebox{1\linewidth}{!}{
\begin{tabular}{lccccc|c}
\toprule
\textbf{Domain} & \multicolumn{1}{c}{\textbf{In-Domain}} & \multicolumn{4}{c|}{\textbf{Out-of-Domain }} & \textbf{ALL}  \\
\cmidrule(lr){1-1} \cmidrule(lr){2-2} \cmidrule(lr){3-6}
\textbf{Method} & \textbf{Life Services} & \textbf{Transportation \& Travel} & \textbf{Healthcare \& Wellness} & \textbf{Social \& Entertainment} & \textbf{Financial Services} & \textbf{AVG} \\
\midrule
Baseline        & \textbf{86.73\%} & 84.50\% & \textbf{90.23\%} & \textbf{87.13\%} & 82.70\% & 86.26\% \\
GRPO            & 29.23\% & 30.00\% & 24.90\% & 25.60\% & 29.37\% & 27.82\% \\
PPO             & 59.20\% & 64.70\% & 62.67\% & 59.70\% & 61.47\% & 61.55\% \\
Reinforce++     & 85.17\% & 76.00\% & 75.63\% & 85.40\% & 83.53\% & 81.15\% \\
Dr.GRPO         & 44.67\% & 49.87\% & 50.17\% & 50.80\% & 46.60\% & 48.42\% \\
\rowcolor{gray!20}\textbf{DFPO (Ours)} & 85.67\% & \textbf{90.50\%} & 89.70\% & 86.40\% & \textbf{84.30\%} & \textbf{87.31\%} \\
\bottomrule
\end{tabular}
}
\label{tab:all-dig}
\end{table*}

\begin{table*}[t]
\centering
\small
\caption{
Accuracy (\%) of different methods trained on the \textbf{Science Domain}.
DFPO achieves the best performance on most benchmarks and consistently leads in ID, OOD, and overall averages, demonstrating strong robustness to noisy supervision and superior generalization under OOD conditions.
}
\resizebox{\columnwidth}{!}{
\begin{tabular}{lccccccc|c|c|c}
\toprule
\textbf{Domain} & \multicolumn{3}{c}{\textbf{In-Domain (Science \& QA)}} & \multicolumn{4}{c|}{\textbf{Out-of-Domain (Math)}} & \textbf{ID} & \textbf{OOD}& \textbf{ALL} \\
\cmidrule(lr){1-1} \cmidrule(lr){2-4} \cmidrule(lr){5-8}
\textbf{Method} & \textbf{SampleQA} & \textbf{GPQA(ALL)} & \textbf{HLE} & \textbf{MATH500} & \textbf{AIME24} & \textbf{Minerva-Math} & \textbf{AMC23} & \textbf{AVG} & \textbf{AVG} & \textbf{AVG}\\
\midrule
Base            & 2.89\% & 3.10\% & 2.89\% & 87.40\% & 41.67\% & 28.68\% & 75.83\% & 2.96\% & 58.40\% & 34.64\%\\
GRPO            & 3.03\% & 2.98\% & 3.24\% & 86.80\% & 45.00\% & 26.84\% & 82.50\% & 3.08\% & 60.29\% & 35.77\% \\
PPO             & 2.82\% & 2.17\% & 3.29\% & 87.00\% & 51.67\% & 29.41\% & 85.00\% & 2.76\% & 63.27\% & 37.34\% \\
Reinforce++     & 3.19\% & 4.35\% & 3.29\% & 89.40\% & 50.00\% & 30.51\% & 83.33\% & 3.61\% & 63.31\% & 37.72\% \\
Dr.\ GRPO       & 2.50\% & 3.99\% & 3.10\% & 88.60\% & 48.33\% & 27.21\% & 80.00\% & 3.20\% & 61.04\% & 36.25\% \\
KTAE            & 3.17\% & 3.99\% & 2.64\% & 87.20\% & 46.67\% & 30.15\% & 82.50\% & 3.27\% & 61.63\% & 36.62\% \\
$\lambda$-GRPO  & 2.77\% & 4.35\% & 3.57\% & 89.60\% & 43.33\% & 29.04\% & 82.50\% & 3.56\% & 61.12\% & 36.45\% \\
BAPO         & 3.10\% & 3.99\% & 3.15\% & 84.40\% & \textbf{60.00\%} & 26.84\% & 82.50\% & 3.41\% & 63.44\% & 37.71\% \\
Robust Bellman  & 3.14\% & 4.17\% & 3.38\% & 87.40\% & 45.00\% & 27.21\% & 83.33\% & 3.56\% & 60.74\% & 36.23\% \\
FlowRL          & \textbf{3.26\%} & 4.17\% & 3.24\% & 89.60\% & 50.00\% & 31.62\% & 86.67\% & 3.56\% & 64.47\% & 38.37\% \\
\rowcolor{gray!20}\textbf{DFPO (Ours)} 
                & 3.17\% & \textbf{5.43\%} & \textbf{4.03\%} 
                & \textbf{91.80\%} & 56.67\% & \textbf{32.35\%} & \textbf{88.33\%} 
                & \textbf{4.21\%} & \textbf{67.29\%} & \textbf{40.25\%} \\
\bottomrule
\end{tabular}
}
\label{tab:all-g}
\end{table*}

\begin{table*}[t]
\centering
\small
\caption{
Accuracy (\%) of different methods trained on the \textbf{Math Domain}.
DFPO achieves the best or tied-best performance across most benchmarks, and attains the highest ID, OOD, and overall averages, demonstrating superior robustness to noisy supervision and improved cross-domain generalization.
}
\resizebox{\columnwidth}{!}{
\begin{tabular}{lccccccc|c|c|c}
\toprule
\textbf{Domain} 
& \multicolumn{4}{c}{\textbf{In-Domain (Math)}} 
& \multicolumn{3}{c|}{\textbf{Out-of-Domain (Science \& QA)}} 
& \textbf{ID} & \textbf{OOD} & \textbf{ALL} \\
\cmidrule(lr){1-1} \cmidrule(lr){2-5} \cmidrule(lr){6-8}
\textbf{Method} 
& \textbf{MATH500} & \textbf{AIME24} & \textbf{Minerva-Math} & \textbf{AMC23} 
& \textbf{SampleQA} & \textbf{GPQA(ALL)} & \textbf{HLE} 
& \textbf{AVG} & \textbf{AVG} & \textbf{AVG} \\
\midrule
Base             & 87.40\% & 41.67\% & 28.68\% & 75.83\% & 2.89\% & 3.10\% & 2.89\% & 58.40\%& 2.96\% & 34.64\%\\
GRPO & 89.20\% & 35.00\% & 28.68\% & 84.17\% & 2.91\% & 4.17\% & 3.57\% & 59.26\% & 3.55\% & 35.39\% \\
PPO & 86.40\% & 36.67\% & 30.51\% & 80.00\% & 2.70\% & 3.99\% & 3.34\% & 58.40\% & 3.34\% & 34.80\% \\
Reinforce++ & 90.20\% & 50.00\% & 30.88\% & 84.17\% & \textbf{3.19\%} & \textbf{4.53\%} & 3.71\% & 63.81\% & 3.81\% & 38.10\% \\
Dr.\ GRPO & 90.80\% & 46.67\% & 30.88\% & \textbf{88.33\%} & 3.12\% & 4.35\% & 3.48\% & 64.17\% & 3.65\% & 38.23\% \\
Robust Bellman & 89.40\% & 36.67\% & \textbf{31.99\%} & 84.17\% & 2.61\% & 3.62\% & 3.43\% & 60.56\% & 3.22\% & 35.98\% \\
KTAE & 86.00\% & 38.33\% & 29.41\% & 78.33\% & 2.91\% & 3.99\% & 3.38\% & 58.02\% & 3.43\% & 34.62\% \\
$\lambda$-GRPO & 90.00\% & 50.00\% & 31.62\% & 80.00\% & 2.91\% & 4.35\% & 3.71\% & 62.91\% & 3.66\% & 37.51\% \\
BAPO & 87.00\% & 38.33\% & 30.51\% & 75.00\% & 3.12\% & 4.35\% & 3.06\% & 57.71\% & 3.51\% & 34.48\% \\
FlowRL & 90.20\% & \textbf{55.00\%} & 29.41\% & 86.67\% & 2.84\% & 3.80\% & 3.56\% & 65.32\% & 3.40\% & 38.78\% \\
\rowcolor{gray!20}\textbf{DFPO (Ours)} & \textbf{91.00\%} & \textbf{55.00\%} & \textbf{31.99\%} & 85.83\% & 2.84\% & \textbf{4.53\%} & \textbf{4.22\%} & \textbf{65.96\%} & \textbf{3.86\%} & \textbf{39.34\%} \\
\bottomrule
\end{tabular}
}
\label{tab:all-m}
\end{table*}

\paragraph{DFPO demonstrates strong stability under noisy real-world feedback}
Tables~\ref{tab:all-dig}, \ref{tab:all-g}, and \ref{tab:all-m} summarize the performance of different methods under noisy and imperfect reward supervision across multiple domains.
On real-world dialogue benchmarks, as shown in Figure~\ref{fig:dialogue_traj}, DFPO continues to improve model performance throughout training and reaches an average accuracy of 86.65\%, without experiencing catastrophic performance collapse.
By contrast, GRPO and Dr.\ GRPO are highly sensitive to reward noise and show substantial performance collapse, with GRPO exhibiting the most severe instability. 
Moreover, DFPO preserves strong performance on reasoning tasks.
After training on Math and Science benchmarks, DFPO achieves overall average accuracies of 39.34\% and 40.25\%, respectively, outperforming GRPO, Reinforce++, and FlowRL, which highlights its robustness under multi-task and multi-domain noisy training.

\paragraph{DFPO maintains consistent effectiveness across in-domain and cross-domain evaluations.}
As reported in Tables~\ref{tab:all-dig}, \ref{tab:all-g}, and \ref{tab:all-m}, DFPO consistently delivers strong in-domain and cross-domain performance.
In dialogue settings, DFPO sustains high task completion rates on previously OOD domains and shows stable performance improvements over the baseline, exceeding the results of Reinforce++ and FlowRL.
Similarly, when trained on either Math or Science tasks, DFPO  consistently achieves the strongest overall performance among all compared methods.
In contrast, robust Bellman PPO exhibit stable convergence, but their generalization to OOD domains remains limited, often producing overly long and incoherent responses.
FlowRL, while improving exploration through reward diffusion, provides limited gains in advantage estimation, which constrains its overall improvement.

\subsection{Ablation Experiment}

\begin{table*}[t]
\centering
\small
\caption{
Ablation of DFPO loss components on scientific tasks. Distributional value modeling with flow matching improves average accuracy from 34.64\% to 38.32\%, while tail and consistency constraints boost OOD accuracy from 64.16\% to 67.12\%. The full method achieves the highest overall accuracy 40.25\%. DCFM denotes Distributional Conditional Flow Matching loss. Consistency Loss denotes Bootstrapped Anchor \& Geometric Consistency Regularization Loss. \textbf{The Standard Distributional Value Flow Modeling PPO} results are under the Above+DCFM Loss methods.
}

\resizebox{\columnwidth}{!}{
\begin{tabular}{lccccccc|c|c|c}
\toprule
\textbf{Domain} & \multicolumn{3}{c}{\textbf{In-Domain (Science \& QA)}} & \multicolumn{4}{c|}{\textbf{Out-of-Domain (Math)}} & \textbf{ID} & \textbf{OOD}& \textbf{ALL} \\
\cmidrule(lr){1-1} \cmidrule(lr){2-4} \cmidrule(lr){5-8}  
\textbf{Loss Component} & \textbf{SampleQA} & \textbf{GPQA(ALL)} & \textbf{HLE} & \textbf{MATH500} & \textbf{AIME24} & \textbf{Minerva-Math} & \textbf{AMC23} & \textbf{AVG} & \textbf{AVG} & \textbf{AVG} \\
\midrule

Base            & 2.89\% & 3.10\% & 2.89\% & 87.40\% & 41.67\% & 28.68\% & 75.83\% & 2.96\% & 58.40\% & 34.64\%\\
Above+Distributional Loss        & 2.94\% & 4.17\% & 3.85\% & 90.80\% & 45.00\% & 30.51\% & 79.17\% & 3.65\% & 61.37\% & 36.63\% \\
Above+DCFM Loss        & 3.19\% & 3.80\% & 3.80\% & 89.60\% & 48.33\% & 31.99\% & 87.50\% & 3.60\% & 64.36\% & 38.32\% \\
Above+Conditional Risk Loss        & 3.10\% & 4.89\% & 3.80\% & 90.40\% & 51.67\% & 31.25\% & 83.33\% & 3.93\% & 64.16\% & 38.35\% \\
Above+Tail Curvature and Shape Loss        & 2.70\% & 4.89\% & 3.66\% & 89.60\% & 51.67\% & 30.88\% & \textbf{88.33\%} & 3.75\% & 65.12\% & 38.82\% \\
Above+Uncertainty DCFM Loss        & 3.10\% & 4.35\% & \textbf{4.08\%} & 89.80\% & 53.33\% & 31.25\% & 86.67\% & 3.84\% & 65.26\% & 38.94\% \\
Above+Consistency Loss        & 3.01\% & 4.53\% & 3.24\% & 90.40\% & \textbf{58.33\%} & \textbf{33.09\%} & 86.67\% & 3.59\% & 67.12\% & 39.90\% \\
\rowcolor{gray!20}\textbf{Above+Lipschitz Optimization(OURS)}        & \textbf{3.17\%} & \textbf{5.43\%} & 4.03\% & \textbf{91.80\%} & 56.67\% & 32.35\% & \textbf{88.33\%} & \textbf{4.21\%} & \textbf{67.29\%} & \textbf{40.25\%} \\
\bottomrule
\end{tabular}
}
\label{tab:ablation-loss-components}
\end{table*}

\begin{table*}[t]
\centering
\small
\caption{Performance comparison on the number of trajectory sampling steps during training.
Single-step sampling achieves the best overall performance, reaching 40.25\% average accuracy, with strong results on MATH500 (91.80\%) and AIME24 (56.67\%).
Increasing the sampling steps to 5 leads to a clear performance drop, while larger steps (10 and 20) partially recover performance on some benchmarks but remain lower overall.}
\resizebox{1\linewidth}{!}{
\begin{tabular}{lccccccc|c|c|c}
\toprule
\textbf{Domain} & \multicolumn{3}{c}{\textbf{In-Domain (Science \& QA)}} & \multicolumn{4}{c|}{\textbf{Out-of-Domain (Math)}} & \textbf{ID} & \textbf{OOD}& \textbf{ALL} \\
\cmidrule(lr){1-1} \cmidrule(lr){2-4} \cmidrule(lr){5-8}  
\textbf{Sampling Steps} & \textbf{SampleQA} & \textbf{GPQA(ALL)} & \textbf{HLE} & \textbf{MATH500} & \textbf{AIME24} & \textbf{Minerva-Math} & \textbf{AMC23} & \textbf{AVG} & \textbf{AVG} & \textbf{AVG} \\
\midrule
Base            & 2.89\% & 3.10\% & 2.89\% & 87.40\% & 41.67\% & 28.68\% & 75.83\% & 2.96\% & 58.40\% & 34.64\%\\
\rowcolor{gray!20}\textbf{1 (Ours)} 
& \textbf{3.17\%} & \textbf{5.43\%} & 4.03\% & \textbf{91.80\%} & \textbf{56.67\%} & \textbf{32.35\%} & 88.33\% 
& \textbf{4.21\%} & \textbf{67.29\%} & \textbf{40.25\%} \\
5 
& 2.87\% & 4.17\% & 3.10\% & 90.20\% & 50.00\% & 31.62\% & 85.83\% 
& 3.38\% & 64.41\% & 38.26\% \\
10 
& 3.05\% & 4.17\% & 3.34\% & 89.40\% & 50.00\% & 31.99\% & 86.67\% 
& 3.52\% & 64.52\% & 38.37\% \\
20 
& 2.98\% & 4.89\% & \textbf{4.22\%} & 90.60\% & 55.00\% & 29.04\% & \textbf{90.00\%} 
& 4.03\% & 66.16\% & 39.53\% \\
\bottomrule
\end{tabular}
}
\label{tab:ablation-sampling}
\end{table*}

\paragraph{Effectiveness of loss components for robust and generalizable value flow learning}
Table~\ref{tab:ablation-loss-components} presents an ablation study of DFPO's loss components on both ID (Science \& QA) and OOD (Math) benchmarks. 
We observe that adding distributional and flow matching losses yields clear gains, raising overall accuracy from 34.64\% to 38.32\%, which confirms that modeling values as flow fields provides a stronger signal under noisy supervision.
The most significant improvements occur with tail-related and flow consistency constraints. For instance, OOD accuracy increases from 64.16\% to 65.12\% with tail-related constraints, and further to 67.12\% with consistency loss. This indicates that constraining value flow trajectories effectively reduces unnecessary exploration and improves robustness in OOD domains.
Our complete model, which integrates these constraints with Lipschitz-based optimization, achieves the highest average accuracy of 40.25\%. This suggests that while Lipschitz optimization helps consolidate structural integrity.

\paragraph{Single-step trajectory sampling is sufficient for stable flow learning}
Table~\ref{tab:ablation-sampling} reports the results with different numbers of trajectory sampling steps during training.
We observe that sampling a single trajectory achieves the best overall performance, indicating that one-step sampling is sufficient to learn the global flow structure of trajectories.
When the sampling number is increased to 5, performance drops noticeably, likely due to sampling bias that increases the impact of noisy or wrong paths.
As the number of sampling steps further increases, performance gradually recovers on some benchmarks, but still does not surpass the single-step setting.
This suggests that while more sampling steps provide richer local flow information, they also make the value model more sensitive to erroneous trajectory regions.
In contrast, single-step sampling captures the overall trajectory distribution while reducing the impact of noisy fluctuations, leading to more stable learning and better generalization.

\subsection{Discussion}

\paragraph{Training flow-based value models is simpler than policy models}

Compared with policy models, training flow-based value models is simpler and more stable. Unlike policy models, value models do not generate language tokens directly; instead, they focus on learning training dynamics and guiding future returns, avoiding sequence generation challenges and reducing sensitivity to linguistic variability. Additionally, value model learning does not rely on importance sampling ratios or explicit action probability estimation, eliminating common approximation errors in policy optimization~\cite{zhu2025llada15variancereducedpreference} and resulting in smaller training variance and more reliable gradient signals. As shown in Figure~\ref{fig:adv-flow}, even a simple flow formulation for value models can produce smooth trajectories that closely follow optimal transport paths, highlighting the practical advantages of applying flow modeling at the value model.

\paragraph{Flow constraints enable fine-grained robustness and generalization}

Flow field modeling enables fine-grained control over training robustness and generalization. As shown in Figure~\ref{fig:adv-vis},~\ref{fig:adv-flow}, DFPO and unconstrained flow methods exhibit similar relative advantage patterns across response tokens. However, unconstrained flow fields tend to have overlapping and collapsing trajectories, reducing exploration and increasing instability. By introducing consistency-based flow constraints, DFPO guides value flow toward smoother, more coherent transport paths, keeping the learned flow closer to optimal transport under noisy supervision. This allows small, controlled trajectory adjustments to enhance value and advantage estimation, enabling precise cross-domain guidance and improved robustness in noisy environments.

\paragraph{Mathematical superiority of time-step flow over scalar regression}
Fundamentally, traditional scalar learning treats value estimation as a regression problem, minimizing prediction error to approximate the conditional expectation $\mathbb{E}[Y|X]$. In complex environments with high variance or multi-modality, this averaging effect leads to \textit{distribution collapse}, where the model outputs a safe mean value that discards critical information regarding risk and diversity~\cite{zhu2025dvpodistributionalvaluemodelingbased}. In contrast, time-step flow modeling learns a continuous evolution process rather than a static target. By modeling a time-dependent vector field $v(z,t)$, the flow defines a diffeomorphism that continuously deforms a simple noise distribution into the complex target distribution via ODE integration~\cite{dong2025valueflows}. This topological flexibility allows the model to preserve high-order statistical moments (e.g., variance and skewness) and faithfully represent disjoint, multi-modal return landscapes that scalar models inherently fail to capture.

\paragraph{Single-step flow inference as a robust global approximation}
While our empirical results favor single-step inference, this approach differs fundamentally from standard single-step direct sampling (e.g., direct diffusion generation or regression). Standard direct sampling often attempts a point-to-point fitting of noisy targets, leading to overfitting of high-frequency reward noise and poor generalization. Conversely, the single-step inference in DFPO represents a global expectation approximation of the entire flow trajectory. By imposing a geometric consistency constraint, we force the vector field to align along straight paths across all time steps $t$. Mathematically, this regularizes the initial velocity $v(z,0)$ to represent the global average velocity pointing toward the target. Consequently, DFPO's single-step projection $z_1 = z_0 + v(z_0, 0)$ functions as a low-pass filter: it effectively captures the global trend of value evolution while filtering out the transient stochastic gradient noise accumulated during multi-step integration. This mechanism renders DFPO significantly more robust in noisy RL post-training regimes compared to standard multi-step solvers.
 \begin{figure}[h]
\centering
\includegraphics[width=0.6 \linewidth]{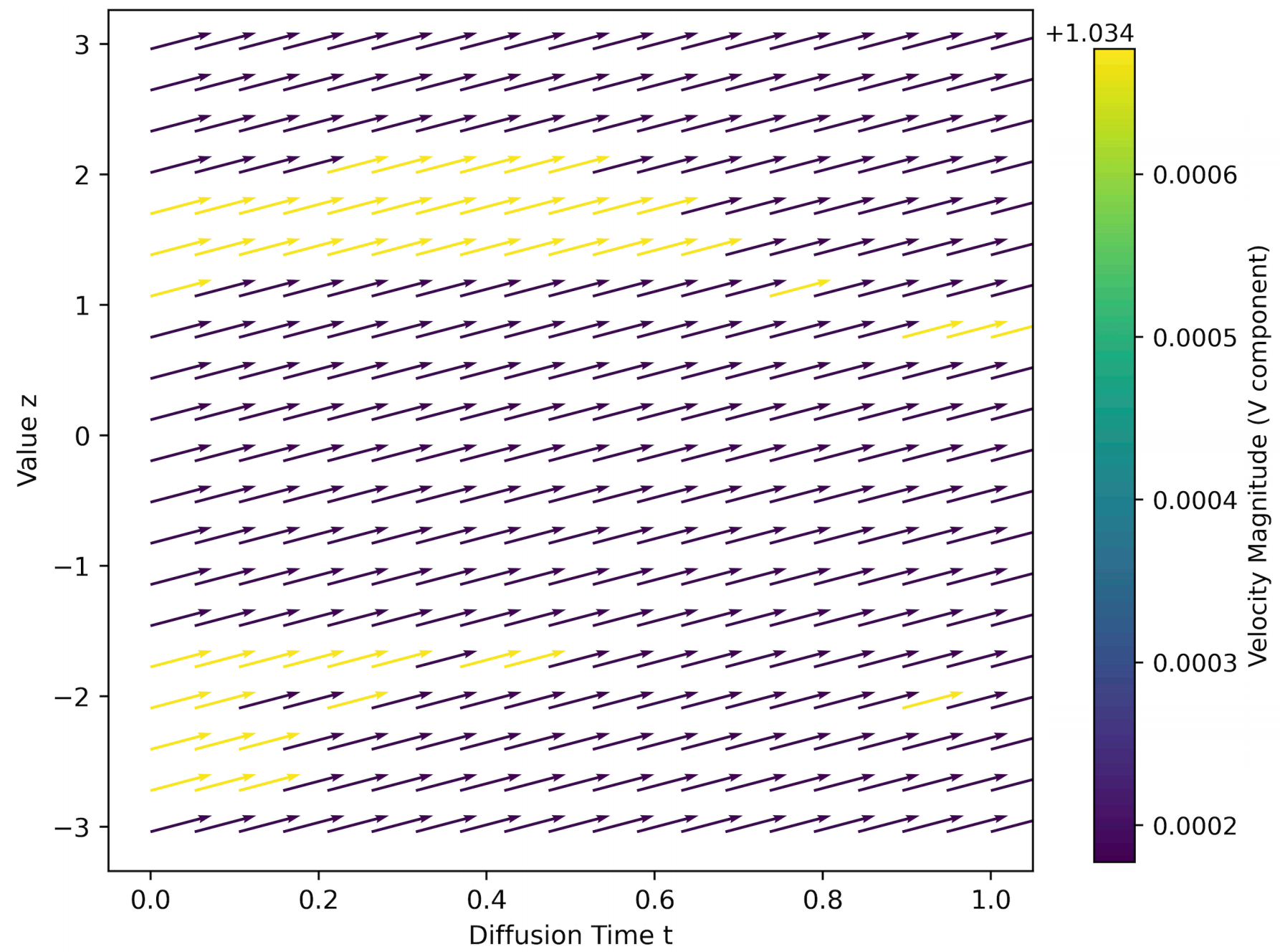}
\caption{Visualization of the learned velocity vector field of one token for the answer part. The model encodes complex distributional structure through the fine-grained modulation of velocity magnitude (indicated by the color scale). As discussed, the vector field dynamically allocates speed based on spatial position, instructing particles in specific regions to accelerate (yellow/green) while neighboring particles decelerate (purple). Through this spatial divergence, the flow effectively compresses, stretches, and splits probability densities to reconstruct complex multi-peak return landscapes at the terminal state.}
\label{fig:velocity_field}
\end{figure}
\paragraph{Modeling complex multi-peak distributions via velocity modulation and Optimal Transport}
A critical question is how straight-line trajectories can theoretically model complex multi-peak return distributions. The mechanism lies in the fine-grained modulation of velocity magnitude within the vector field. As shown in Figure~\ref{fig:velocity_field}, while trajectories are geometrically rectified to be straight, the vector field $v(z,t)$ dynamically allocates speed based on spatial position, for instance, instructing particles in the center to accelerate leftward while neighboring particles decelerate. Through this spatial divergence, the flow effectively compresses, stretches, and splits probability densities to reconstruct complex multi-peak landscapes.~\cite{lipman2023flowmatchinggenerativemodeling,chen2019neuralordinarydifferentialequations}
Furthermore, we enforce Optimal Transport (OT) paths to ensure non-crossing and parallel dynamics. In ODE theory, crossing trajectories imply conflicting velocity directions at a single point, leading to numerical collapse; OT naturally selects a ``non-entangled'' transport plan (e.g., mapping left-side noise to left-side targets)~\cite{dupont2019augmentedneuralodes}, yielding a smooth, continuous vector field that is inherently robust. Mathematically, straight paths minimize the Wasserstein transport cost (energy)~\cite{liu2022flowstraightfastlearning}, representing the simplest constant-velocity function for neural networks to approximate. This imposes a strong regularization that prevents the model from wasting capacity on chaotic, winding paths, allowing it to focus entirely on capturing the distribution's shape. Ultimately, this balance between constraint and expressivity serves as the bedrock for exploration: by ensuring that the predicted multi-peak structure and variance reflect true environmental uncertainty rather than approximation artifacts, the model provides high-fidelity signals for risk-aware decision-making.

\paragraph{Token-level value modeling ensures robust adherence to formatting constraints}

The formatting errors observed in Robust Bellman PPO, BAPO, and FlowRL during real-world dialogue evaluation can be attributed to their reliance on coarse-grained value supervision. Robust Bellman PPO optimizes the lower bound of scalar global returns, which provides a trajectory-level signal but lacks the token-level resolution necessary to enforce strict structural formatting. Similarly, BAPO applies global policy clipping without fine-grained regularization across the token sequence, making it susceptible to overlooking structural tokens under high noise. FlowRL models action-space reward distributions globally rather than tracking token-level value trajectories, resulting in supervision that struggles to stably guide the generation of specific formatting markers. In contrast, DFPO models continuous distributional value flows at the token level. This token-level supervision, combined with consistency constraints, provides the fine-grained guidance required to maintain adherence to structural and formatting constraints under noisy and out-of-domain conditions.

\section{Conclusion}

We introduce DFPO, a robust distributional RL framework that models value as continuous temporal flows. It improves robustness and generalization under noisy supervision by learning value flow fields instead of discrete scalar quantiles, further scaling the value model to guide and enhance the performance in real-world environments. By integrating conditional risk control and consistency constraints along value flow trajectories, DFPO effectively balances stable optimization with exploratory efficiency. Extensive experiments on dialogue, math, and scientific tasks show DFPO consistently outperforms PPO, FlowRL, and other robust baselines, providing a scalable, practical solution for real-world RL and LLM post-training.

\section*{Impact Statement}

While DFPO demonstrates strong robustness and generalization under noisy supervision and OOD scenarios, several limitations remain.
First, the optimal configuration of the composite objective function, specifically the balance between flow matching terms ($\lambda_{\text{reg}}, \lambda_{\text{cons}}$) and distributional risk constraints ($\lambda_{\text{risk}}, \lambda_{\text{shape}}$) may vary across different domains, requiring task-specific tuning to achieve the best trade-off between exploration and stability.
Second, although the 1-step Euler solver significantly accelerates inference, the assumption of straight-line trajectories relies heavily on the convergence of the geometric consistency loss; in scenarios with extremely highly non-linear value dynamics, the approximation error might increase.
Finally, while DFPO improves stability under moderate noise via confidence weighting and anchor regularization, extreme reward corruption or completely noisy supervision can still degrade performance, as the flow field may eventually collapse to erroneous attractors. This paper presents work whose goal is to advance the field of Machine
Learning. There are many potential societal consequences of our work, none
which we feel must be specifically highlighted here.

\bibliography{dfpo}

\newpage
\appendix
\begin{algorithm}[H]
\caption{DFPO Training Algorithm}
\label{alg:dfpo}
\begin{algorithmic}[1]
\small
\REQUIRE Dataset $\mathcal{D}$, Policy $\pi_\theta$, Value Model $V_\phi$ with Flow Head $v_\phi$.
\REQUIRE Hyperparameters: $\lambda_{\text{reg}}, \lambda_{\text{cons}}, \lambda_{\text{risk}}, \lambda_{\text{shape}}$, quantile levels $\alpha, \beta$.
\ENSURE Optimized Policy $\pi_\theta$ and Value Model $V_\phi$.

\FOR{each iteration}
    \STATE \textbf{Sample Trajectories:} Generate trajectories $\tau = \{(s_t, a_t, r_t, s_{t+1})\}$ using current policy $\pi_\theta$.

    \STATE \textbf{Distributional Value Estimation (Inference):}
    \FOR{each state $s_t$ in trajectories}
        \STATE Extract hidden state: $\mathbf{h}_{\mathbf{s}} = \text{Transformer}(s_t)$
        \STATE Sample initial noise batch: $\{z_0^{(k)}\}_{k=1}^K \sim \mathcal{N}(0, 1)$
        \STATE Solve ODE IVP (Euler): $\{z_1^{(k)}\} = \text{ODESolve}(v_\phi, z_0^{(k)}, t=0 \to 1, \mathbf{h}_{\mathbf{s}})$
        \STATE Obtain predicted value distribution quantiles: $\hat{Z}(s_t) = \text{Sort}(\{z_1^{(k)}\})$
    \ENDFOR

    \STATE \textbf{Compute Distributional GAE:}
    \FOR{each timestep $t$ reversed}
        \STATE Calculate distributional TD error (element-wise on sorted supports):
        \STATE $\delta_{D, t} = R_t + \gamma \hat{Z}(s_{t+1}) - \hat{Z}(s_t)$
        \STATE Recursive Distributional Advantage:
        \STATE $A_D(s_t) = \delta_{D, t} + (\gamma \lambda) A_D(s_{t+1})$
        \STATE Compute Target Returns for Value Model: $Z^{\text{tgt}}(s_t) = \hat{Z}(s_t) + A_D(s_t)$
    \ENDFOR

    \STATE \textbf{Value Model Update:}
    \FOR{each mini-batch $B$ sampled from buffer}
        \STATE \textbf{1. Flow Matching Objectives:}
        \STATE Sample time $t \sim \mathcal{U}[0, 1]$, noise $x_0 \sim \mathcal{N}(0, 1)$, target sample $x_1 \sim Z^{\text{tgt}}(s)$
        \STATE Compute Confidence Weight: $w_{\text{conf}}(\mathbf{s}) = \sigma(\|J(1)\|^2 / \tau_{\text{temp}}) + 0.5$
        \STATE Construct optimal path: $z_t = t x_1 + (1-t) x_0$
        \STATE Anchor target: $x_1^{\text{anc}} = \text{StopGrad}(\mathbb{E}[x_1|\mathbf{s}])$, path $z_t^{\text{anc}} = t x_1^{\text{anc}} + (1-t) x_0$
        \STATE $\mathcal{L}_{\text{UDCFM}} = \mathbb{E}_{t, x_0, x_1} [ w_{\text{conf}}(\mathbf{s}) \| v_\phi(z_t, t, \mathbf{h}_{\mathbf{s}}) - (x_1 - x_0) \|^2 ]$
        \STATE $\mathcal{L}_{\text{BCFM}} = \mathbb{E}_{t, x_0} [ w_{\text{conf}}(\mathbf{s}) \| v_\phi(z_t^{\text{anc}}, t, \mathbf{h}_{\mathbf{s}}) - (x_1^{\text{anc}} - x_0) \|^2 ]$

        \STATE \textbf{2. Geometric Consistency Regularization:}
        \STATE Sample $t \sim \mathcal{U}[0, 1]$, set conjugate $t' = 1-t$
        \STATE Project terminals: $\hat{x}_1(z_t) = z_t + (1-t)v_\phi(z_t)$, \quad $\hat{x}_1(z_{t'}) = z_{t'} + (1-t')v_\phi(z_{t'})$
        \STATE $\mathcal{L}_{\text{cons}} = \| \hat{x}_1(z_t) - \hat{x}_1(z_{t'}) \|^2$

        \STATE \textbf{3. Risk-Controlled Distributional Constraints:}
        \STATE Sort current predictions $\hat{z}$ and targets $z^{\text{tgt}}$
        \STATE Calculate Risk Loss (Left/Right Tails):
        \STATE $\mathcal{L}_{\text{Risk}} = \| \text{Tail}_{L}(\hat{z}) - \text{Tail}_{L}(z^{\text{tgt}}) \|^2 + \| \text{Tail}_{R}(\hat{z}) - \text{Tail}_{R}(z^{\text{tgt}}) \|^2$
        \STATE Calculate Shape Loss (Curvature Penalty):
        \STATE $\mathcal{L}_{\text{Shape}} = \sum_{k \in \mathcal{I}_L} [\nabla^2 \hat{z}_{(k)}]_+ + \sum_{k \in \mathcal{I}_R} [-\nabla^2 \hat{z}_{(k)}]_+$

        \STATE \textbf{Total Critic Loss:}
        \STATE $\mathcal{L}_{\text{Total}} = (\mathcal{L}_{\text{UDCFM}} + \lambda_{\text{reg}}\mathcal{L}_{\text{BCFM}} + \lambda_{\text{cons}}\mathcal{L}_{\text{cons}}) + (\lambda_{\text{risk}}\mathcal{L}_{\text{Risk}} + \lambda_{\text{shape}}\mathcal{L}_{\text{Shape}})$
        \STATE Update $\phi$ using $\nabla_\phi \mathcal{L}_{\text{Total}}$
    \ENDFOR

    \STATE \textbf{Policy Update:}
    \STATE Scalarize Advantage for PPO: $\hat{A}_t = \text{Mean}(A_D(s_t))$
    \STATE $\mathcal{L}_{\text{PPO}} = \mathbb{E}[\min(\rho_t \hat{A}_t, \text{clip}(\rho_t, 1-\epsilon, 1+\epsilon)\hat{A}_t)]$
    \STATE Update $\theta$ using $\nabla_\theta \mathcal{L}_{\text{PPO}}$
\ENDFOR
\end{algorithmic}
\end{algorithm}

\section{Additional Details for DFPO}
\label{sec:appendix}
\subsection{Pseudocode}
The full algorithm of DFPO is detailed in Algorithm ~\ref{alg:dfpo}.

\subsection{Efficient Jacobian Estimation via Sensitivity ODE}
\label{app:jacobian_ode}

To quantify flow uncertainty efficiently, we approximate the sensitivity of the terminal state $z_1$ with respect to the initial noise $z_0$ without relying on computationally expensive Monte Carlo sampling.~\cite{dong2025valueflows} We employ the adjoint sensitivity method to propagate the local Jacobian $J(t) := \frac{\partial z_t}{\partial z_0}$ along the flow trajectory.
Differentiating the flow equation $\frac{dz_t}{dt} = v_\theta(z_t, t, \mathbf{h}_{\mathbf{s}})$, the evolution of the Jacobian follows the linear ODE:
\begin{equation}
    \frac{dJ(t)}{dt} = \frac{\partial v_\theta(z_t, t, \mathbf{h}_{\mathbf{s}})}{\partial z_t} J(t), \quad \text{with } J(0) = 1.
\end{equation}
In our implementation, we solve this coupled system $(z_t, J_t)$ in a single forward pass using a fixed-step Euler solver. Specifically, we set the number of ODE solver steps to $N=10$.
It is crucial to note that $N=10$ represents the temporal discretization resolution of the numerical integration, not the number of random samples.
At each step $t_k$, we compute the scalar gradient $\nabla_z v_\theta$ via automatic differentiation and update the Jacobian: $J_{t_{k+1}} \approx J_{t_k} + (\nabla_z v_\theta \cdot J_{t_k}) \Delta t$.
Since we utilize a 1D value distribution, this process incurs only negligible computational overhead (computing a single scalar gradient per step) compared to the Transformer backbone, while providing a stable deterministic estimate of flow sensitivity.

\section{Theoretical Analysis}
\label{sec:theory}

In this section, we provide the theoretical justification for the convergence of DFPO.

\subsection{Contraction of the Distributional GAE Operator}
\label{proof:contraction}

\textbf{Proposition 1.}
\textit{
Let $\mathcal{T}^\pi$ denote the distributional Bellman operator.
The Distributional GAE operator $\mathcal{T}^\pi_{\mathrm{GAE}}$, defined as a geometric mixture of $k$-step distributional Bellman operators, is a contraction mapping under the supremum $1$-Wasserstein metric for any $\gamma < 1$.
Consequently, the iterative application of $\mathcal{T}^\pi_{\mathrm{GAE}}$ converges to a unique fixed-point return distribution.
}

\begin{proof}
Let $\mathcal{Z}$ denote the space of return distributions with finite first moments.
We equip this space with the supremum $1$-Wasserstein metric (also known as the Earth Mover's Distance):
\begin{equation}\small
    \bar d_1(Z_1, Z_2) := \sup_{s,a} W_1\!\left(Z_1(s,a), Z_2(s,a)\right).
\end{equation}

First, consider the standard one-step distributional Bellman operator $\mathcal{T}^\pi$, defined as:
\begin{equation}\small
    \mathcal{T}^\pi Z(s,a)
    \stackrel{D}{=}
    R(s,a) + \gamma Z(s',a'), \qquad a' \sim \pi(\cdot \mid s').
\end{equation}
Leveraging the translation invariance ($W_1(X+c, Y+c) = W_1(X, Y)$) and absolute homogeneity ($W_1(cX, cY) = |c|W_1(X, Y)$) of the Wasserstein metric, and assuming a synchronous coupling of transitions and actions, we have:
\begin{equation}\small
\begin{aligned}
    W_1\!\left(\mathcal{T}^\pi Z_1(s,a), \mathcal{T}^\pi Z_2(s,a)\right)
    &= \gamma \, W_1\!\left(Z_1(s',a'), Z_2(s',a')\right) \\
    &\le \gamma \, \bar d_1(Z_1, Z_2).
\end{aligned}
\end{equation}
Taking the supremum over $(s,a)$ yields $\bar d_1(\mathcal{T}^\pi Z_1, \mathcal{T}^\pi Z_2) \le \gamma \, \bar d_1(Z_1, Z_2)$, establishing that $\mathcal{T}^\pi$ is a $\gamma$-contraction.
By induction, the $k$-step operator $(\mathcal{T}^\pi)^k$ is a $\gamma^k$-contraction.

Next, consider the Distributional GAE operator $\mathcal{T}^\pi_{\mathrm{GAE}}$. It is constructed as a geometric mixture of $k$-step returns, parameterized by $\lambda \in [0,1]$:
\begin{equation}\small
    \mathcal{T}^\pi_{\mathrm{GAE}} Z
    \stackrel{D}{=}
    \sum_{k=1}^{\infty} (1-\lambda)\lambda^{k-1} \, (\mathcal{T}^\pi)^k Z.
\end{equation}
A key property of the $1$-Wasserstein metric is its convexity with respect to mixture distributions. For any set of distributions $\{\mu_k, \nu_k\}$ and weights $w_k$ summing to 1, $W_1(\sum w_k \mu_k, \sum w_k \nu_k) \le \sum w_k W_1(\mu_k, \nu_k)$.
Applying this property to the GAE operator:
\begin{equation}\small
\begin{aligned}
    &\bar d_1(\mathcal{T}^\pi_{\mathrm{GAE}} Z_1, \mathcal{T}^\pi_{\mathrm{GAE}} Z_2)\\
    &\le \sum_{k=1}^{\infty} (1-\lambda)\lambda^{k-1} \,
    \bar d_1\!\left((\mathcal{T}^\pi)^k Z_1, (\mathcal{T}^\pi)^k Z_2\right) \\
    &\le (1-\lambda) \sum_{k=1}^{\infty} \lambda^{k-1} \gamma^k \, \bar d_1(Z_1, Z_2) \\
    &= \gamma(1-\lambda) \left( \sum_{j=0}^{\infty} (\lambda\gamma)^j \right) \bar d_1(Z_1, Z_2) \\
    &= \frac{\gamma(1-\lambda)}{1-\lambda\gamma} \, \bar d_1(Z_1, Z_2).
\end{aligned}
\end{equation}

Let $\Gamma := \frac{\gamma(1-\lambda)}{1-\lambda\gamma}$.
Since $\gamma < 1$ and $\lambda \in [0,1]$, it follows that $\Gamma < 1$. Thus, $\mathcal{T}^\pi_{\mathrm{GAE}}$ is a contraction mapping with factor $\Gamma$.
By the Banach Fixed-Point Theorem, repeated application of $\mathcal{T}^\pi_{\mathrm{GAE}}$ converges to a unique fixed-point return distribution.
\end{proof}

\subsection{Sensitivity-Aware Flow Optimization}
\label{proof:sensitivity}

\textbf{Proposition 2.} \textit{The Uncertainty-Weighted Distributional CFM (UDCFM) objective enhances the learning of complex flow dynamics by prioritizing regions with high flow sensitivity, thereby preventing over-smoothing in state-dependent transport paths.}

\begin{proof}
Consider the regression target in Flow Matching, which aims to fit the vector field $v_\theta$ to the target velocity $u_t = x_1 - x_0$. In real-world RL, the supervision is often multi-modal or sparse, leading to regions where the optimal transport path exhibits high complexity.

Let $\mathcal{R}_{\text{complex}}$ denote the subset of the state space where the flow dynamics exhibit high sensitivity to initial conditions, characterized by a large Jacobian norm $\|J(1)\|^2$.
Standard Flow Matching optimization minimizes the expected MSE: $\mathcal{L}_{\text{FM}} = \mathbb{E}[\| v_\theta - u_t \|^2]$.
Under limited capacity or noisy supervision, the minimizer of this unweighted objective tends to converge to the conditional mean $\mathbb{E}[x_1 | x_0]$, which often results in \textit{over-smoothed} vector fields that fail to capture the sharp transitions in $\mathcal{R}_{\text{complex}}$.

Our UDCFM objective introduces a sensitivity-dependent weight:
\begin{equation}\small
    w_{\text{conf}}(\mathbf{s}) \propto \sigma(\|J(1)\|^2 / \tau)
\end{equation}
Since the sigmoid function is monotonically increasing, this weighting scheme assigns higher importance to samples with large Jacobian norms (i.e., high-sensitivity regions).
Instead of treating high sensitivity purely as aleatoric noise to be discarded, we interpret it as a signal of topological complexity or hard examples that require stronger gradient updates to be learned correctly.

By assigning $w_{\text{conf}}(\mathbf{s}) > 1$ to samples in $\mathcal{R}_{\text{complex}}$, the gradient descent process focuses more capacity on minimizing the approximation error in these difficult regions:
\begin{equation}\small
    \nabla \mathcal{L}_{\text{UDCFM}} \approx \mathbb{E} \left[ (1 + \alpha \|J(1)\|^2) \nabla \| v_\theta - u_t \|^2 \right]
\end{equation}
This acts as a form of curriculum learning or focal loss, ensuring that the learned policy does not collapse to a trivial mean solution in high-dynamic scenarios, thus improving the robustness of the value distribution's structural integrity.
\end{proof}

\subsection{Geometric Consistency Implies Straight Trajectories}
\label{proof:consistency}

\textbf{Proposition 3.} \textit{If the geometric consistency loss $\mathcal{L}_{\text{cons}}$ is minimized to zero for all $t \in [0, 1)$, then the learned vector field $v_\theta$ along the flow trajectory is constant in time, implying that the generated flow trajectories are straight lines.}

\begin{proof}
Let $z_t$ be the state at time $t$ governed by the ODE $d z_t / dt = v_\theta(z_t, t)$.
The geometric consistency constraint enforces that the projected terminal state $\hat{x}_1(z_t)$ is invariant with respect to time $t$. The projection operator is defined based on a linear flow assumption:
\begin{equation}\small
    \hat{x}_1(z_t, t) = z_t + (1-t) v_\theta(z_t, t)
\end{equation}
Minimizing $\mathcal{L}_{\text{cons}}$ to zero implies that $\hat{x}_1(z_t, t)$ is constant for all $t$ along the trajectory. Taking the total time derivative of $\hat{x}_1$ and setting it to zero:
\begin{equation}\small
    \frac{d}{dt} \hat{x}_1(z_t, t) = \frac{d z_t}{dt} + \frac{d}{dt} \left[ (1-t) v_\theta(z_t, t) \right] = 0
\end{equation}
Substituting $d z_t / dt = v_\theta(z_t, t)$ and applying the product rule:
\begin{equation}\small
\begin{aligned}
    v_\theta(z_t, t) + \left[ (-1) \cdot v_\theta(z_t, t) + (1-t) \frac{d}{dt} v_\theta(z_t, t) \right] &= 0 \\
    v_\theta(z_t, t) - v_\theta(z_t, t) + (1-t) \frac{d v_\theta}{dt} &= 0 \\
    (1-t) \frac{d v_\theta}{dt} &= 0
\end{aligned}
\end{equation}
For any $t < 1$, this implies that the total derivative of the velocity field along the trajectory is zero:
\begin{equation}\small
    \frac{d v_\theta(z_t, t)}{dt} = 0 \implies v_\theta(z_t, t) = \mathbf{c} \quad (\text{constant vector})
\end{equation}
Consequently, the trajectory $z_t$ is the integral of a constant velocity, which is a straight line:
\begin{equation}\small
    z_t = z_0 + \int_0^t \mathbf{c} \, d\tau = z_0 + t \cdot \mathbf{c}
\end{equation}
This proves that enforcing geometric consistency is theoretically equivalent to regularizing the flow towards a straight-line Optimal Transport path.
\end{proof}

\subsection{Sufficiency of 1-Step Euler Solver}
\label{proof:onestep}

\textbf{Proposition 4.} \textit{Under the assumption that the flow matching objective and consistency regularization are optimized, the learned vector field approximates a straight-line trajectory. In this regime, the numerical error of a 1-Step Euler solver vanishes, rendering it sufficient for high-precision inference.}

\begin{proof}
Our training objective combines Conditional Flow Matching with Geometric Consistency. As shown in Proposition 3, the consistency term drives the vector field $v_\theta$ to be constant in time, i.e., $v_\theta(z_t, t) \approx v_0$ for all $t \in [0, 1]$.

Consider the numerical integration of the ODE $d z_t = v_\theta(z_t, t) dt$ from $t=0$ to $t=1$. The exact solution is:
\begin{equation}\small
    z_1 = z_0 + \int_0^1 v_\theta(z_t, t) dt
\end{equation}
A 1-Step Euler solver approximates this integral as:
\begin{equation}\small
    \hat{z}_1^{\text{Euler}} = z_0 + (1 - 0) \cdot v_\theta(z_0, 0)
\end{equation}
The local truncation error of the Euler method is bounded by $\mathcal{O}(\Delta t^2 \max_t \| \ddot{z}_t \|)$. Since $\ddot{z}_t = \frac{d}{dt} v_\theta(z_t, t)$, and Proposition 3 implies $\frac{d v_\theta}{dt} \to 0$, the acceleration term vanishes ($\ddot{z}_t \approx 0$).

Therefore, as the flow trajectory becomes a straight line (Straight Flow), the Euler approximation becomes exact:
\begin{equation}\small
    \int_0^1 v_\theta(z_t, t) dt = \int_0^1 v_0 dt = v_0 = v_\theta(z_0, 0)
\end{equation}
This theoretical result aligns with the findings of Rectified Flow~\cite{liu2022flowstraightfastlearning}, which demonstrates that minimizing the transport cost yields straight displacement paths. Our method effectively performs a "One-Step Rectification" during training by enforcing the target $x_1 - x_0$ (straight path) and the consistency constraint, allowing the model to generate high-fidelity samples with a single function evaluation (NFE=1) during inference, avoiding the computational cost of iterative solvers used in diffusion models.
\end{proof}

\section{Additional Experimental Details}
\label{appendix:setup}

\subsection{Data Collection and Preprocessing}
\paragraph{Math and Scientific Reasoning}
For mathematical reasoning tasks, we utilize Qwen3-8B as the base model to generate pseudo-labels from 39,000 samples in the Light-R1 dataset \cite{wen2025lightr1curriculumsftdpo}. A filtering process based on 5 rounds of majority voting is applied: only those samples where at least 3 out of 5 responses are identical are retained. This results in a final dataset of 31,209 instances suitable for RL training.
For scientific QA tasks, pseudo-labels are derived from 26,529 samples in the SuperGPQA dataset. Applying the same 5-round voting mechanism but with a lower threshold (at least 2 consistent responses), we preserve 10,075 samples for training. To prevent data leakage and ensure a fair assessment of generalization, the evaluation benchmarks are strictly disjoint from the training data.

\paragraph{Real-World Dialogue}
We utilize Honor-Dialogue, a dataset derived from authentic real-world interactions. The construction of Honor-Dialogue follows a goal-oriented, scenario-centric paradigm. We initially defined the agent's role as a pragmatic, context-aware assistant, imposing rigorous constraints regarding accuracy, naturalness, and adherence to service norms. We then curated  many representative domains, one general and lots of professional domains, based on high-frequency real-life interactions. Each domain is defined by explicit triggers, such as user intent and contextual features, alongside standardized protocols for response logic and information prioritization.
For every domain, we synthesized realistic user inputs comprising current messages and dialogue history, paired with standardized outputs aligned with specific goals. To ensure high-quality supervision, we explicitly annotated dialogue states, response content, and intended targets. A representative sample is shown in Figure ~\ref{fig:dialogue-example-insurance}, where sensitive details (e.g., user IDs) are masked with "xxx" to strictly adhere to data privacy regulations.

 For the reward model, training involves 36,000 labeled instances, with a subset of 3,000 held out for validation. The policy model undergoes fine-tuning and subsequent reinforcement learning on a separate corpus of 50,952 conversations; notably, this training phase excludes data from the financial, social entertainment, transportation, and healthcare sectors.
To evaluate performance, we test across five distinct scenarios: Daily Services (1,004 samples), Financial Services (700), Social Entertainment (534), Transportation (463), and Healthcare (380). By encompassing a wide array of practical tasks, this dataset mirrors real-world complexities. Crucially, this level of multi-category, real-scenario diversity is absent in contemporary datasets.

\subsection{Dialogue Evaluation Methods}

\begin{table*}[t]
\centering
\caption{
Statistics of the validation dataset across 11 real-world service scenarios.
The dataset features diverse interaction types with varying dialogue lengths, providing a robust testbed for evaluating generalization.
}
\small
\resizebox{1\linewidth}{!}{
\begin{tabular}{lccccccccccc|c}
\toprule
\textbf{Scenario} & Wealth & Rental & Insurance & Food & Express & Promotion & Loan & Housing & Service & Product & General & Avg \\
\midrule
Dialogue Count & 87 & 99 & 138 & 120 & 215 & 66 & 70 & 87 & 67 & 69 & 92 & 94.73 \\
Avg Turns      & 5.40 & 4.22 & 5.22 & 3.37 & 3.76 & 4.55 & 4.33 & 4.68 & 5.46 & 4.44 & 3.46 & 4.44 \\
\bottomrule
\end{tabular}
}
\label{tab:validation-data-statistics}
\end{table*}

Our evaluation employs GPT-4o strictly as a rubric-based evaluator rather than a subjective judge. To ensure reproducibility, the model is constrained by a prompt containing precise metrics, a 1–5 grading scale, scoring examples, and formatting requirements. We confirmed reliability via \textbf{1)} stability testing through multiple runs and \textbf{2)} validation against human scores. Specifically, we conducted a comparative study using 1,110 samples across 11 scenarios, the statistics of which appear in Table \ref{tab:validation-data-statistics}. Four professional annotators scored this data using identical guidelines.
\begin{table*}[ht]
\centering
\small
\caption{Agreement analysis between human annotators and GPT-4o. 
The average scores are extremely close, indicating high reliability of the automated evaluation.}
\resizebox{0.7\linewidth}{!}{
\begin{tabular}{lcccccc|cc}
\toprule
\textbf{Category} & \multicolumn{4}{c}{\textbf{Human Annotators}} & \multicolumn{2}{c|}{\textbf{GPT Models}} & \textbf{Human AVG} & \textbf{Model AVG}  \\
\cmidrule(lr){1-1} \cmidrule(lr){2-5} \cmidrule(lr){6-7} \cmidrule(lr){8-9}
Score & 4.595  & 4.689  & 4.490  & 4.330  & 4.510  & 4.501  & 4.526     & 4.505     \\
\bottomrule
\end{tabular}
}
\label{tab:human-model-agreement}
\end{table*}
As presented in Table \ref{tab:human-model-agreement}, the results indicate near-perfect alignment with a marginal mean score difference of 0.021 between human annotators and GPT-4o. These findings confirm a strong consistency between human and automated scores, thereby validating the effectiveness of our rubric. While we acknowledge that establishing full external validity and reproducibility requires further independent annotation, we are in the process of releasing our rubric, evaluation prompts, and the real-world dialogue dataset to facilitate community replication. To illustrate our assessment logic, Figure \ref{fig:content-relevance-guidelines} displays the core prompt section used for evaluating response content relevance.

\subsection{Additional experimental details in the Dialogue Task}

\begin{table*}[t]
\centering
\small
\caption{
Comprehensive results on the Real-world Dialogue Benchmark at 1000 training steps under noisy reward supervision. We report Task Completion (TC), Ask Completion (AC), and Goal Completion (GC) rates for each domain. DomainAVG (D-AVG) represents the average performance across these three metrics within a domain, while AVG denotes the overall average across all five domains. DFPO achieves the highest overall average, consistently outperforming all prior RL baselines in both in-domain and out-of-domain settings.
}
\resizebox{1\linewidth}{!}{
\begin{tabular}{lcccc|cccc|cccc|cccc|cccc|c}
\toprule
\multirow{2}{*}{\textbf{Method}} &
\multicolumn{4}{c|}{\textbf{Life Services (In-Domain)}} &
\multicolumn{4}{c|}{\textbf{Transportation \& Travel}} &
\multicolumn{4}{c|}{\textbf{Healthcare \& Wellness}} &
\multicolumn{4}{c|}{\textbf{Social \& Entertainment}} &
\multicolumn{4}{c|}{\textbf{Financial Services}} &
\multirow{2}{*}{\textbf{AVG}} \\
\cmidrule(lr){2-21}
& TC & AC & GC & D-AVG
& TC & AC & GC & D-AVG
& TC & AC & GC & D-AVG
& TC & AC & GC & D-AVG
& TC & AC & GC & D-AVG
& \\
\midrule
Baseline & \textbf{72.5} & \textbf{95.0} & \textbf{92.7} & \textbf{86.7} & 66.7 & \textbf{96.5} & 90.3 & 84.5 & \textbf{77.2} & 97.8 & \textbf{95.7} & \textbf{90.2} & \textbf{75.3} & 92.0 & \textbf{94.1} & \textbf{87.1} & 65.7 & 91.4 & \textbf{91.0} & 82.7 & 86.3 \\
GRPO & 47.4 & 14.6 & 25.7 & 29.2 & 41.0 & 20.5 & 28.5 & 30.0 & 24.6 & 21.6 & 28.5 & 24.9 & 43.3 & 10.0 & 23.5 & 25.6 & 50.9 & 12.7 & 24.5 & 29.4 & 27.8 \\
PPO & 43.8 & 63.4 & 70.4 & 59.2 & 46.2 & 77.1 & 70.8 & 64.7 & 31.6 & 77.4 & 79.0 & 62.7 & 47.4 & 61.5 & 70.2 & 59.7 & 45.3 & 70.4 & 68.7 & 61.5 & 61.5 \\
Reinforce++ & 70.9 & 92.5 & 92.1 & 85.2 & 54.1 & 87.9 & 86.0 & 76.0 & 45.8 & 94.3 & 86.8 & 75.6 & 66.1 & \textbf{99.2} & 90.9 & 85.4 & \textbf{69.6} & 92.4 & 88.6 & 83.5 & 81.1 \\
Dr.GRPO & 34.5 & 45.3 & 54.2 & 44.7 & 37.0 & 53.0 & 60.0 & 49.9 & 23.0 & 66.0 & 62.0 & 50.2 & 44.0 & 50.0 & 58.0 & 50.8 & 35.0 & 54.0 & 51.0 & 46.6 & 48.4 \\
\rowcolor{gray!20}
\textbf{DFPO (Ours)} & 70.8 & 94.4 & 91.8 & 85.7 & \textbf{82.1} & 94.5 & \textbf{94.9} & \textbf{90.5} & 75.4 & \textbf{98.8} & 94.9 & 89.7 & 71.1 & 96.8 & 91.3 & 86.4 & 69.4 & \textbf{92.6} & 90.9 & \textbf{84.3} & \textbf{87.3} \\
\bottomrule
\end{tabular}
}
\label{tab:dialog-full}
\end{table*}

\begin{table*}[t]
\caption{
Performance comparison on the effect of risk interval control weight in DFPO.
A moderate interval weight of 0.1 leads to the best balance between stability and generalization across in-domain (Science \& QA) and out-of-domain (Math) benchmarks.
}
\centering
\small
\resizebox{1\linewidth}{!}{
\begin{tabular}{lccccccc|c|c|c}
\toprule
\textbf{Domain} & \multicolumn{3}{c}{\textbf{In-Domain (Science \& QA)}} & \multicolumn{4}{c|}{\textbf{Out-of-Domain (Math)}} & \textbf{ID} & \textbf{OOD} & \textbf{ALL} \\
\cmidrule(lr){1-1} \cmidrule(lr){2-4} \cmidrule(lr){5-8}
\textbf{Risk Interval} & \textbf{SampleQA} & \textbf{GPQA(ALL)} & \textbf{HLE} 
 & \textbf{MATH500} & \textbf{AIME24} & \textbf{Minerva-Math} & \textbf{AMC23} 
 & \textbf{AVG} & \textbf{AVG} & \textbf{AVG} \\
\midrule
Base            & 2.89\% & 3.10\% & 2.89\% & 87.40\% & 41.67\% & 28.68\% & 75.83\% & 2.96\% & 58.40\% & 34.64\%\\

0    & 3.01\% & 4.17\% & 3.52\% & 89.20\% & 50.00\% & 30.51\% & 83.33\% & 3.57\% & 63.26\% & 37.68\% \\

0.05 & 3.21\% & 5.07\% & \textbf{4.08\%} & 90.60\% & 55.00\% & 31.99\% & 84.17\% & 4.12\% & 65.44\% & 39.16\% \\

\rowcolor{gray!20}\textbf{0.1 (Ours)} 
     & 3.17\% & \textbf{5.43\%} & 4.03\% 
     & \textbf{91.80\%} & \textbf{56.67\%} & \textbf{32.35\%} & \textbf{88.33\%} 
     & \textbf{4.21\%} & \textbf{67.29\%} & \textbf{40.25\%} \\

0.2  & \textbf{3.31\%} & 4.71\% & 3.38\% & 89.20\% & 46.67\% & 29.04\% & 84.17\% & 3.80\% & 62.27\% & 37.21\% \\
\bottomrule
\end{tabular}
}
\label{tab:ablation-fx}
\end{table*}

\begin{table*}[t]
\caption{
Performance comparison on a Qwen3-1.7B model under noisy supervision.
DFPO consistently achieves the best overall performance 33.75\%, demonstrating strong scalability and robustness across domains.
}
\centering
\small
\resizebox{1\linewidth}{!}{
\begin{tabular}{lccccccc|c|c|c}
\toprule
\textbf{Domain} & \multicolumn{3}{c}{\textbf{In-Domain (Science \& QA)}} & \multicolumn{4}{c|}{\textbf{Out-of-Domain (Math)}} & \textbf{ID} & \textbf{OOD}& \textbf{ALL} \\
\cmidrule(lr){1-1} \cmidrule(lr){2-4} \cmidrule(lr){5-8}  
\textbf{Method} & \textbf{SampleQA} & \textbf{GPQA(ALL)} & \textbf{HLE} & \textbf{MATH500} & \textbf{AIME24} & \textbf{Minerva-Math} & \textbf{AMC23} & \textbf{AVG} & \textbf{AVG} & \textbf{AVG} \\
\midrule
Base & 1.78\% & 2.36\% & 2.87\% & 84.80\% & 35.00\% & 19.49\% & 69.17\% & 2.34\% & 52.12\% & 30.78\% \\
GRPO            & 1.60\% & 1.99\% & 3.94\% & 82.20\% & 33.33\% & 21.32\% & 77.50\% & 2.51\% & 53.59\% & 31.70\% \\
PPO             & \textbf{1.91\%} & 2.54\% & 3.94\% & 83.80\% & 33.33\% & 22.43\% & \textbf{79.17\%} & 2.80\% & 54.68\% & 32.45\% \\
Reinforce++     & 1.48\% & 2.36\% & 3.52\% & 84.60\% & 38.33\% & 23.16\% & 75.83\% & 2.45\% & 55.48\% & 32.75\% \\
Dr.\ GRPO       & 1.50\% & 2.17\% & 3.10\% & 84.00\% & 35.00\% & 22.79\% & 76.67\% & 2.26\% & 54.62\% & 32.18\% \\
KTAE            & 1.64\% & \textbf{2.90\%} & \textbf{4.36\%} & 82.20\% & 26.67\% & 23.90\% & 64.17\% & \textbf{2.97\%} & 49.24\% & 29.41\% \\
$\lambda$-GRPO  & 1.62\% & \textbf{2.90\%} & 3.89\% & \textbf{85.00\%} & \textbf{40.00\%} & 22.06\% & 70.83\% & 2.80\% & 54.47\% & 32.33\% \\
BAPO          & 1.55\% & 2.36\% & 3.48\% & 82.40\% & 30.00\% & 19.85\% & 64.17\% & 2.46\% & 49.11\% & 29.12\% \\
Robust Bellman  & 1.64\% & 2.17\% & 4.17\% & 84.60\% & 31.67\% & 22.79\% & 75.00\% & 2.66\% & 53.52\% & 31.72\% \\
FlowRL          & 1.76\% & 1.81\% & 3.99\% & 84.00\% & 36.67\% & 22.06\% & 75.83\% & 2.52\% & 54.64\% & 32.30\% \\
\rowcolor{gray!20}\textbf{DFPO (Ours)}
                & 1.76\% & 2.54\% & 3.99\%
                & \textbf{85.00\%} & \textbf{40.00\%} & \textbf{24.63\%} & 78.33\%
                & 2.76\% & \textbf{56.99\%} & \textbf{33.75\%} \\
\bottomrule
\end{tabular}
}
\label{tab:small-model}
\end{table*}

\begin{table*}[t]
\centering
\small
\caption{
Performance comparison on the weight of the flow consistency loss.
A moderate setting (0.01) achieves the best overall performance, reaching the highest average accuracy of 40.25\%.
Stronger or weaker consistency constraints lead to lower performance, especially on OOD benchmarks.
}
\resizebox{1\linewidth}{!}{
\begin{tabular}{lccccccc|c|c|c}
\toprule
\textbf{Domain} & \multicolumn{3}{c}{\textbf{In-Domain (Science \& QA)}} & \multicolumn{4}{c|}{\textbf{Out-of-Domain (Math)}} & \textbf{ID} & \textbf{OOD}& \textbf{ALL} \\
\cmidrule(lr){1-1} \cmidrule(lr){2-4} \cmidrule(lr){5-8}  
\textbf{Consistency Weight} & \textbf{SampleQA} & \textbf{GPQA(ALL)} & \textbf{HLE} & \textbf{MATH500} & \textbf{AIME24} & \textbf{Minerva-Math} & \textbf{AMC23} & \textbf{AVG} & \textbf{AVG} & \textbf{AVG} \\
\midrule
Base            & 2.89\% & 3.10\% & 2.89\% & 87.40\% & 41.67\% & 28.68\% & 75.83\% & 2.96\% & 58.40\% & 34.64\%\\
0.001       & \textbf{3.17\%} & 3.99\% & 3.75\% & 90.00\% & 48.33\% & 31.25\% & 80.83\% & 3.64\% & 62.60\% & 37.33\% \\
0.005       & 2.96\% & 4.35\% & 3.71\% & 90.60\% & \textbf{60.00\%} & 29.78\% & \textbf{88.33\%} & 3.67\% & 67.18\% & 39.96\% \\
\rowcolor{gray!20}\textbf{0.01 (Ours)} & \textbf{3.17\%} & \textbf{5.43\%} & \textbf{4.03\%} & \textbf{91.80\%} & 56.67\% & \textbf{32.35\%} & \textbf{88.33\%} & \textbf{4.21\%} & \textbf{67.29\%} & \textbf{40.25\%} \\
0.05        & 3.01\% & 3.99\% & 3.61\% & 89.60\% & 51.67\% & 31.25\% & 83.33\% & 3.54\% & 63.96\% & 38.07\% \\
\bottomrule
\end{tabular}
}
\label{tab:ablation-consistency}
\end{table*}

\paragraph{DFPO establishes superior final convergence stability}
Table~\ref{tab:dialog-full} presents the final performance across all five dialogue domains. While standard RL baselines suffer from severe instability, DFPO achieves the highest overall final average of 86.7\%, significantly outperforming Reinforce++ (82.8\%) and PPO (60.1\%).
Notably, PPO and GRPO-based methods exhibit a drastic performance collapse at the end of training, with PPO dropping to near 60\% averages in most domains and GRPO falling below 30\%.
This validates our core motivation: without the continuous, risk-controlled flow field, scalar value models are prone to learning unstable policy updates that eventually diverge. In contrast, DFPO maintains robust convergence, matching or exceeding the strongest baselines even in the final training stages.

\paragraph{DFPO prevents catastrophic forgetting and oscillation}
Figure~\ref{fig:dialogue_traj} shows the training dynamics, and integrating the trajectory analysis from Table~\ref{tab:dialogue-results} with the final results in Table~\ref{tab:dialog-full} reveals a critical advantage in learning dynamics.
Baselines like PPO and Reinforce++ exhibit a "peak-and-degrade" pattern: they reach competitive performance at intermediate steps, such as Step 250, but suffer from catastrophic forgetting as training proceeds. For instance, PPO degrades from an early peak of 88.8\% to a final 60.1\%.
DFPO, however, exhibits a flat and stable performance curve, sustaining an average of $\sim$86.6\% throughout the entire training process. This empirically proves that the continuous, risk-controlled flow field learned by DFPO effectively regularizes the learning path, preventing the policy from collapsing into erroneous attractors induced by noisy supervision.

\paragraph{Robust generalization across diverse domains}
At the 1000 training steps, DFPO demonstrates remarkable resilience in OOD settings where other methods falter.
In the Healthcare \& Wellness domain, DFPO achieves an average of 89.7\%, significantly surpassing Reinforce++ (75.6\%) and PPO (62.7\%). Similarly, in the Transportation \& Travel domain, DFPO maintains 90.5\%, whereas PPO collapses to 64.7\%.
Even in domains with highly complex semantics, such as the Social \& Entertainment domain, DFPO retains high precision (86.4\%), largely outperforming Dr.GRPO (50.8\%).
These results confirm that the continuous, risk-controlled flow field learned by DFPO captures stable semantic features that generalize well, ensuring reliable performance even when baseline methods fail to converge.

\subsection{Additional Ablation Experiment}

\paragraph{Proper risk interval control improves robustness and generalization.}
Table~\ref{tab:ablation-fx} illustrates the effect of different risk interval settings in DFPO.
When risk interval control is removed and the setting is 0, the value model becomes more sensitive to noisy supervision and shows weaker and less stable performance across tasks.
When the interval constraint is further increased to a strong level, overall performance drops on most benchmarks, suggesting that excessive restriction limits effective exploration.
Overall, the moderate setting 0.1 adopted in DFPO achieves the best balance between stability and exploration, reaching the highest average accuracy of 40.25\%.
\paragraph{DFPO scales effectively across different model sizes}
Table~\ref{tab:small-model} reports results on a smaller-scale model under noisy supervision.
Despite the reduced model capacity, DFPO consistently outperforms PPO, Reinforce++, and FlowRL.
In particular, DFPO achieves the highest out-of-domain average accuracy of 56.99\% on math tasks and maintains competitive performance on science benchmarks.
These results show that DFPO remains stable and effective even on smaller models, indicating good scalability and robust generalization under noisy training conditions.

\paragraph{Proper consistency constraints balance stability and flexibility in flow learning}
Table~\ref{tab:ablation-consistency} shows the effect of different flow consistency loss weights during training.
A proper consistency constraint helps the model reach a good balance between constrained and divergent flow learning.
With a moderate weight of 0.01, DFPO achieves the best overall performance, reaching an average accuracy of 40.25\%, which is clearly higher than the stronger setting of 0.05 (38.07\%).
When the consistency weight is too large, the flow becomes overly constrained, making value guidance conservative and limiting generalization to out-of-domain states.
In contrast, when the weight is too small, the flow tends to diverge, and noisy trajectories introduce instability during training.
These results show that an appropriate consistency constraint is important for stable optimization and robust generalization.

\paragraph{DFPO demonstrates strong statistical stability across multiple training seeds}
To evaluate statistical significance and training stability, we conducted additional experiments using multiple training random seeds on the scientific reasoning tasks. 
As shown in Table~\ref{tab:multi-seed}, DFPO exhibits strong robustness against initialization and sampling variance. 
The overall average accuracy shows only minor fluctuations, consistently remaining within a tight and high-performing range of 39.74\% to 40.25\%. 
Notably, on complex out-of-domain mathematical benchmarks such as MATH500 and AIME24, the model's performance remains highly stable, maintaining scores strictly above 90.60\% and 55.00\%, respectively, regardless of the seed setting. 
These results confirm that the superior performance of DFPO is statistically robust and deeply rooted in the method's algorithmic design rather than random chance.

\begin{table*}[t]
\centering
\small
\caption{
Performance comparison of DFPO on the reasoning tasks across multiple random seeds. 
The results demonstrate that DFPO maintains stable and superior performance regardless of seed variance, with only minor fluctuations across both ID and OOD benchmarks.
}
\resizebox{1\linewidth}{!}{
\begin{tabular}{lccccccc|c|c|c}
\toprule
\textbf{Domain} & \multicolumn{3}{c}{\textbf{In-Domain (Science \& QA)}} & \multicolumn{4}{c|}{\textbf{Out-of-Domain (Math)}} & \textbf{ID} & \textbf{OOD}& \textbf{ALL} \\
\cmidrule(lr){1-1} \cmidrule(lr){2-4} \cmidrule(lr){5-8}  
\textbf{Seed} & \textbf{SampleQA} & \textbf{GPQA(ALL)} & \textbf{HLE} & \textbf{MATH500} & \textbf{AIME24} & \textbf{Minerva-Math} & \textbf{AMC23} & \textbf{AVG} & \textbf{AVG} & \textbf{AVG} \\
\midrule
22          & 3.05\% & 4.83\% & \textbf{4.26\%} & 90.80\% & 55.00\% & \textbf{32.72\%} & 87.50\% & 4.05\% & 66.51\% & 39.74\% \\
32          & 3.10\% & 5.25\% & 4.03\% & 91.40\% & 55.00\% & 31.99\% & 87.50\% & 4.13\% & 66.47\% & 39.75\% \\
\rowcolor{gray!20}\textbf{42 (Ours)} & 3.17\% & \textbf{5.43\%} & 4.03\% & \textbf{91.80\%} & \textbf{56.67\%} & 32.35\% & 88.33\% & \textbf{4.21\%} & 67.29\% & \textbf{40.25\%} \\
52          & \textbf{3.21\%} & 4.53\% & 3.66\% & 90.60\% & \textbf{56.67\%} & 31.99\% & \textbf{90.00\%} & 3.80\% & \textbf{67.32\%} & 40.09\% \\
\bottomrule
\end{tabular}
}
\label{tab:multi-seed}
\end{table*}

\section{Additional Visualization}
\begin{figure*}[h]
    \centering
    \includegraphics[width=0.9\textwidth]{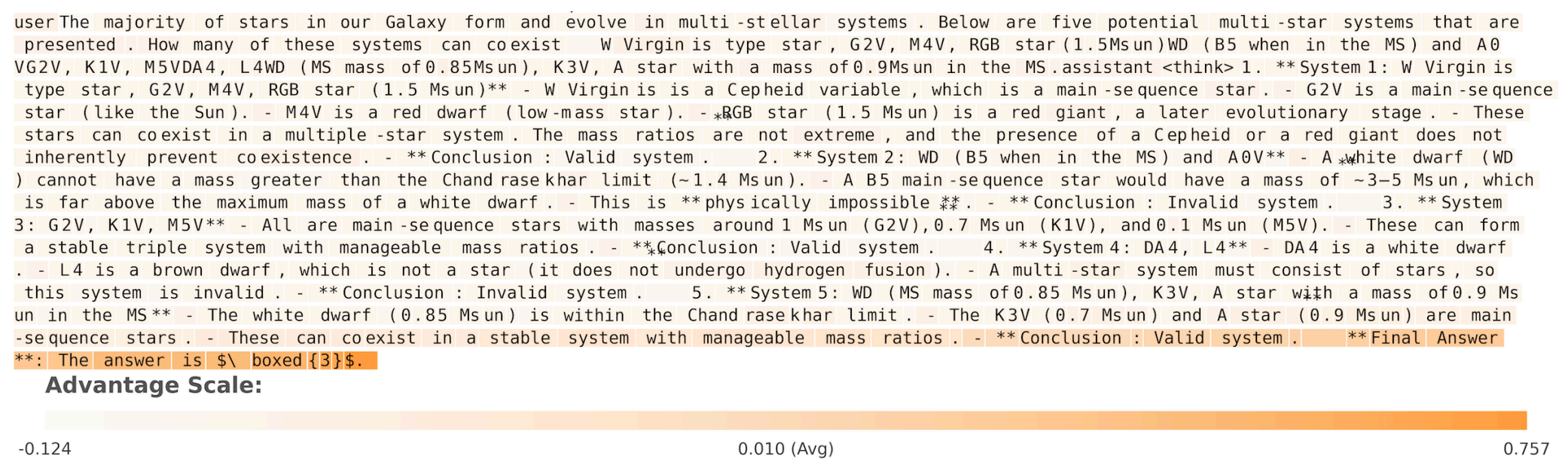}
    \caption{\textbf{Advantage estimation visualization of the PPO method.} The method fails to effectively highlight or capture key semantic tokens within the generated response.}
    \label{fig:advppo}
\end{figure*}

\begin{figure*}[h]
    \centering
    \includegraphics[width=0.9\textwidth]{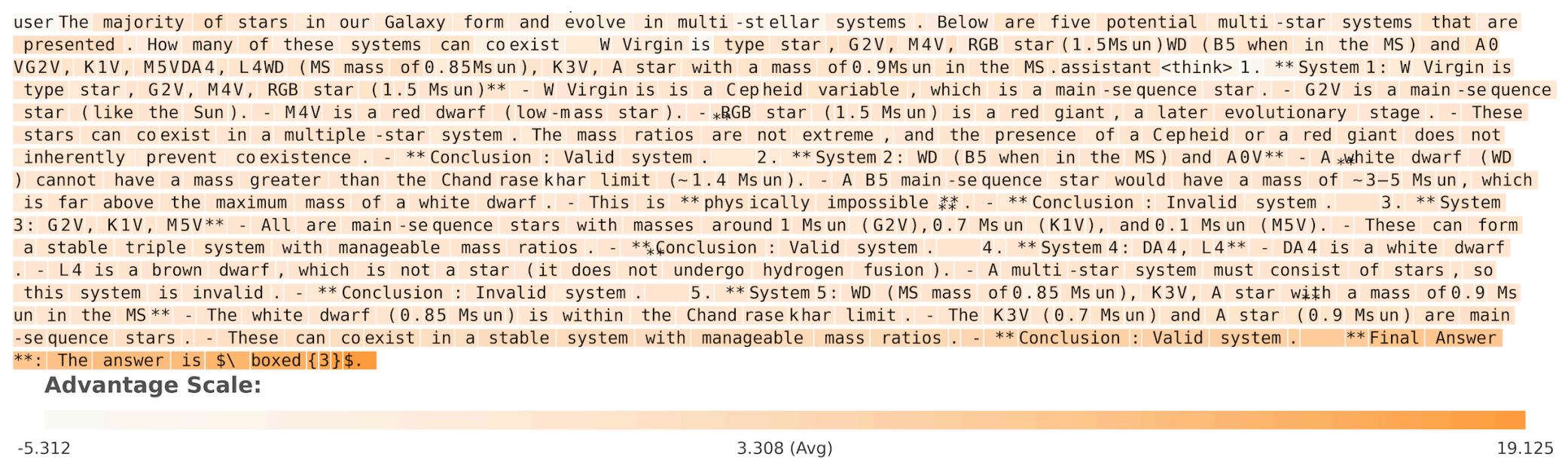}
    \caption{\textbf{Advantage estimation visualization of the Robust Bellman PPO method.} The method demonstrates significant instability and fails to localize key semantic terms. It exhibits extreme fluctuations in advantage values, with peak overestimation reaching as high as 19.125.}
    \label{fig:advbei}
\end{figure*}

\begin{figure*}[h]
    \centering
    \includegraphics[width=0.9\textwidth]{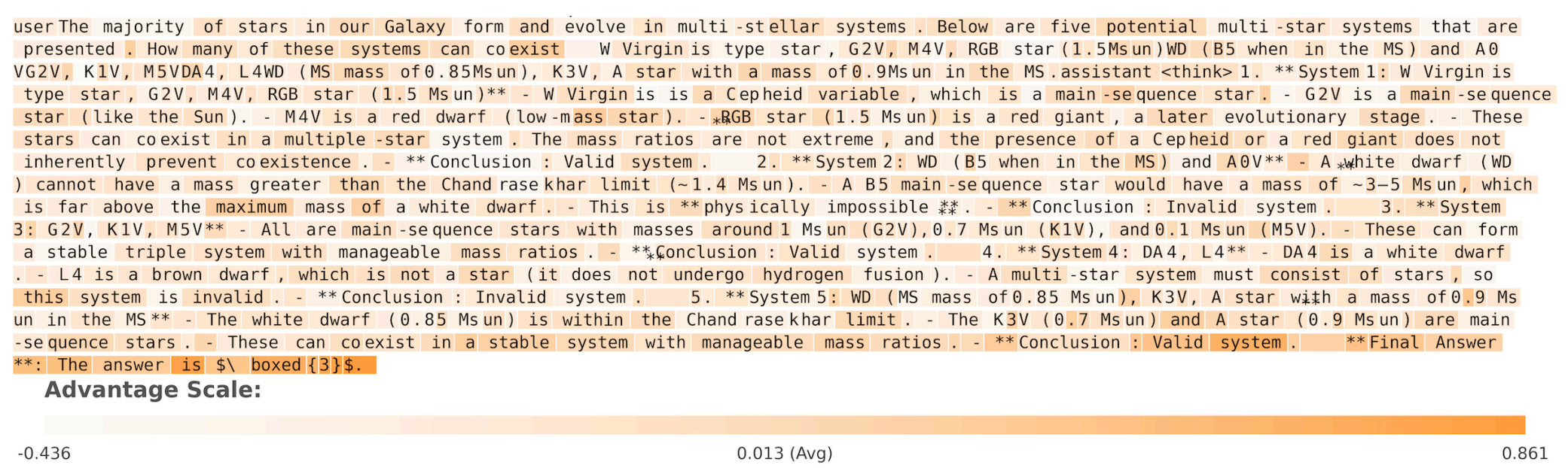}
    \caption{\textbf{Advantage estimation visualization of the Standard Distributional Value Flow Modeling PPO method.} While this method successfully identifies key words, it lacks fine-grained control over the numerical magnitude of advantage estimates, resulting in suboptimal value precision.}
    \label{fig:advflow}
\end{figure*}

\begin{figure*}[h]
    \centering
    \includegraphics[width=0.9\textwidth]{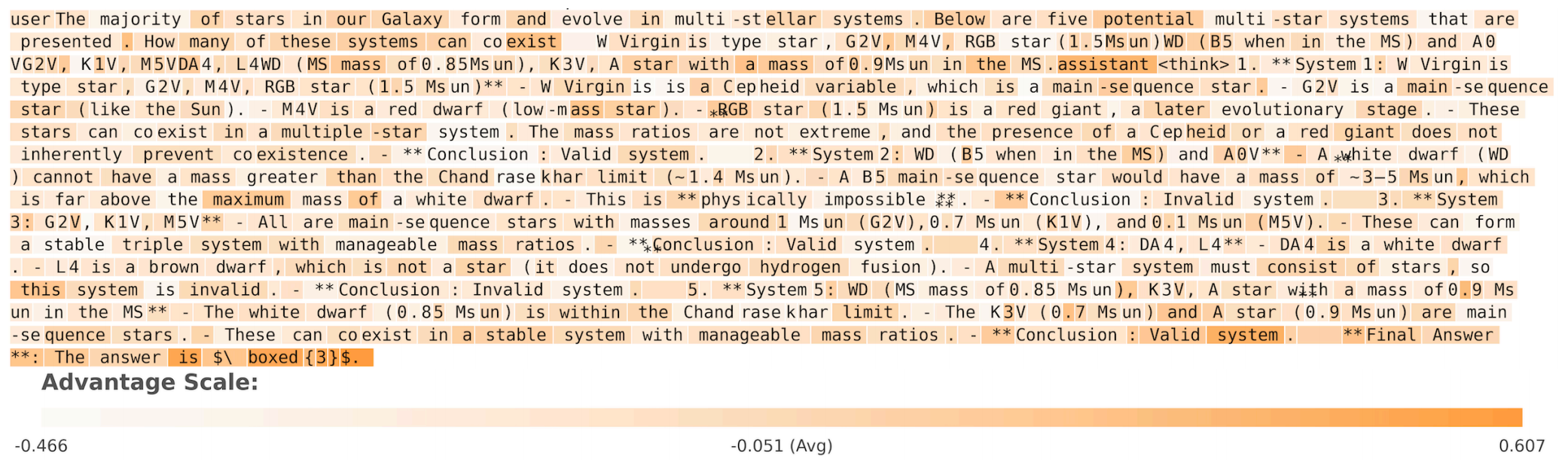}
    \caption{\textbf{Advantage estimation visualization of our proposed DFPO method.} The method accurately captures key semantic terms (e.g., "maximum" and "mass") via distinct advantage weights while simultaneously achieving fine-grained control over the value details, demonstrating superior stability under noisy supervision.}
    \label{fig:advour}
\end{figure*}

Figures~\ref{fig:advppo}, \ref{fig:advbei}, \ref{fig:advflow}, and \ref{fig:advour} illustrate a comparative analysis of advantage estimations among PPO, Robust Bellman PPO, Standard Distributional Value Flow Modeling PPO, and our proposed DFPO method for the same output sequence.
Notably, significant discrepancies in advantage estimation patterns and semantic sensitivity are observed across the four methods.
For PPO (Figure~\ref{fig:advppo}), the method fails to capture key semantic information, treating most tokens with uniform importance, which reflects its limited capability in distinguishing critical decision points.
Similarly, the Robust Bellman PPO method (Figure~\ref{fig:advbei}) also fails to recognize key semantic tokens. Furthermore, it demonstrates severe instability in value estimation, characterized by excessive fluctuations and significant overestimation, with advantage values peaking at an abnormal high of 19.125. 
The Distributional Value Flow Modeling PPO method (Figure~\ref{fig:advflow}) shows improvement by successfully identifying and highlighting key words. However, it falls short in the precise regulation of advantage values, lacking fine-grained control over the numerical details, which may lead to suboptimal policy updates.
In contrast, our DFPO method (Figure~\ref{fig:advour}) demonstrates superior performance. It not only accurately captures key semantic terms (e.g., "maximum" and "mass") by assigning distinct and interpretable advantage weights but also achieves fine-grained control over the advantage values. This balance ensures that the advantage signals are both semantically meaningful and numerically stable, validating the effectiveness of our proposed risk-controlled flow modeling.

\begin{table}[h]
\centering
\small
\caption{Trajectory-level generalization performance in the dialogue task across different methods and training steps.  We report Task Completion Rates(TCR), Ask Completion Rates(ACR), and Goal Completion Rates(GCR) for each methods. Our method maintains high performance even in last stages, beyond GRPO, PPO, Reinforce++, and Dr.GRPO.}
\begin{tabular}{lcccc}
\toprule
\textbf{Step} & \textbf{TCR} & \textbf{ACR} & \textbf{GCR} & \textbf{AVG} \\
\midrule
\multicolumn{5}{l}{\textit{Cold Start}} \\
0 & 71.5\% & 94.5\% & 92.8\% & 86.3\% \\ 
\midrule
\multicolumn{5}{l}{\textit{GRPO}} \\
250 & 35.6\% & \textbf{28.2\%} & \textbf{34.9\%} & \textbf{32.8\%} \\
500 & 41.2\% & 16.5\% & 28.5\% & 28.8\% \\
1000 & 41.4\% & 15.9\% & 26.1\% & 27.8\% \\
\rowcolor{gray!30}\textbf{Last} & \textbf{43.1\%} & 15.1\% & 26.9\% & 28.4\% \\
\midrule
\multicolumn{5}{l}{\textit{PPO}} \\
250 & \textbf{75.8\%} & \textbf{96.9\%} & \textbf{93.9\%} & \textbf{88.8\%} \\
500 & 71.8\% & 93.3\% & 90.5\% & 85.2\% \\
750 & 49.7\% & 78.5\% & 76.8\% & 68.4\% \\
1000 & 42.9\% & 70.0\% & 71.8\% & 61.5\% \\
\rowcolor{gray!30}\textbf{Last} & 42.1\% & 67.7\% & 70.4\% & 60.1\% \\
\midrule
\multicolumn{5}{l}{\textit{Reinforce++}} \\
250 & \textbf{72.9\%} & \textbf{97.7\%} & \textbf{92.8\%} & \textbf{87.8\%} \\
500 & 63.9\% & 96.7\% & 90.9\% & 83.9\% \\
750 & 64.2\% & 95.1\% & 89.6\% & 83.0\% \\
1000 & 61.3\% & 93.3\% & 88.9\% & 81.1\% \\
\rowcolor{gray!30}\textbf{Last} & 63.7\% & 94.9\% & 89.9\% & 82.8\% \\
\midrule
\multicolumn{5}{l}{\textit{Dr.GRPO}} \\
250 & \textbf{60.2\%} & \textbf{82.2\%} & \textbf{85.0\%} & \textbf{75.9\%} \\
500 & 38.1\% & 62.3\% & 68.8\% & 56.3\% \\
750 & 33.5\% & 55.5\% & 60.4\% & 49.8\% \\
1000 & 34.6\% & 53.7\% & 56.9\% & 48.4\% \\
\rowcolor{gray!30}\textbf{Last} & 34.1\% & 51.5\% & 54.9\% & 46.9\% \\
\midrule
\multicolumn{5}{l}{\textbf{\textit{OURS}}} \\
250 & 72.6\% & 94.9\% & 92.3\% & 86.6\% \\
500 & 71.4\% & \textbf{95.7\%} & 92.4\% & 86.5\% \\
750 & 72.4\% & 94.3\% & \textbf{92.8\%} & 86.5\% \\
1000 & \textbf{73.8\%} & 95.4\% & \textbf{92.8\%} & \textbf{87.3\%} \\
\rowcolor{gray!30}\textbf{Last} & 73.0\% & 94.3\% & 92.6\% & 86.7\% \\
\bottomrule
\end{tabular}
\label{tab:dialogue-results}
\end{table}

\begin{figure*}[!t]
\centering

\begin{tcolorbox}[
  colback=gray!4!white,
  colframe=gray!85!black,
  title=\textbf{System Prompt},
  fonttitle=\bfseries
]
\textbf{Role}: You are the AI call assistant of xxx Inc., capable of answering phone calls clearly and politely on behalf of the user. 
Based on predefined dialogue goals, you should first identify the appropriate goal according to the caller’s intent and then respond accordingly.
You do not have access to any personal information about the user beyond the task description, and you must inform the caller that any provided information will be conveyed to the user.

\textbf{Dialogue Goals (Customized)}:
\begin{itemize}
  \item \textbf{0: General Domain} – Used when the call does not belong to a specific domain. Maintain politeness and professionalism, clarify the caller’s intent.
  \item \textbf{1: Medical \& Health (Treatment Adjustment)} – Express concern about medication changes, inquire about alternatives and scientific evidence, and ask about reports or prescription updates.
  \item \textbf{2: Life Services} – If service completion is reported, express dissatisfaction, request compensation, and ask about service details and response time.
  \item \textbf{3: Medical \& Health (Insurance Policy Update)} – Express satisfaction with policy changes, notify the user, raise concerns about reimbursement of specific medications, and inquire about the conditions and follow-up services of newly covered dental examinations.
\end{itemize}
\end{tcolorbox}

\begin{tcolorbox}[
  colback=gray!4!white,
  colframe=gray!85!black,
  title=\textbf{Caller History and New Input},
  fonttitle=\bfseries,
  boxrule=1pt,
  top=3mm,
  bottom=3mm
]
\textbf{Conversation History}:

\textit{Caller}: Hello, this is the Municipal Medical Insurance Service Center. 
We would like to inform you that there have been recent adjustments to the medical insurance policy, including the removal of a certain application process and the addition of dental examinations to the reimbursement list. 
These changes will officially take effect at the end of this month. 
Please also note that some specific surgical procedures will no longer be covered.

\textit{Assistant}: Hello, this call is currently handled by the AI assistant on behalf of the user. 
May I ask whether there are any changes to the reimbursement policy for specific medications?

\vspace{1mm}
\textbf{Caller New Input}: Don’t worry, there are no changes to the reimbursement policy for specific medications. 
The adjustments mainly concern surgical procedures. 
Medication-related updates were already covered in last month’s notice.
\end{tcolorbox}

\begin{tcolorbox}[
  colback=teal!2!white, 
  colframe=teal!80!black, 
  coltitle=white,
  title=\textbf{Model Response},
  fonttitle=\bfseries
]
\textbf{Target Domain}: \textit{3: Medical \& Health (Insurance Policy Update)}

\textbf{Dialogue State}: \textit{Ongoing}

\textbf{Response}: Thank you for the clarification. 
Could you please provide more details about the specific conditions and follow-up services for the newly covered dental examination reimbursement?
\end{tcolorbox}

\caption{A multi-turn example from the Honor-Dialogue dataset under the medical insurance policy update scenario. 
The model correctly identifies the dialogue goal and maintains an ongoing conversation state while proactively requesting detailed conditions for newly introduced reimbursement items.}
\label{fig:dialogue-example-insurance}
\end{figure*}

\begin{figure*}[!t]
\centering

\begin{tcolorbox}[
  colback=gray!4!white,
  colframe=gray!85!black,
  title=\textbf{Core Prompt for Response Content Relevance},
  fonttitle=\bfseries,
  boxrule=1pt,
  top=3mm,
  bottom=3mm
]
This task evaluates the content relevance of the callee’s latest response in an incoming call scenario.
The callee is expected to communicate clearly and naturally with the caller while actively collecting useful information.

When scoring the callee’s latest response, please follow the steps below:

\textbf{Step 1.} Read and understand the full dialogue history and the callee’s latest response.

\textbf{Step 2.} Score from the dimension of content relevance.  
Content relevance is defined as whether the callee’s latest response is directly related to the caller’s statements or questions, and whether it provides a reasonable and context-aware reply based on the caller’s input.
Use a 1–5 scoring scale with the following criteria:

\begin{itemize}
    \item \textbf{1 point:} The response is completely irrelevant to the caller’s input and dialogue context. It fails to address the caller’s statement or question and provides unrelated or meaningless information.

    \item \textbf{2 points:} The response is weakly related. Although it stays within the same general scenario, it contains noticeable irrelevant content or fails to properly respond to the caller’s main intent.

    \item \textbf{3 points:} The response is partially relevant. It aligns with the general scenario, but the callee misunderstands the caller’s key intent, shifts focus away from the main issue, or guides the caller toward an action that is reasonable but beyond the scope of the current conversation.

    \item \textbf{4 points:} The response is mostly relevant. It addresses the caller’s input and provides additional useful information, though the suggested action or guidance may not fully comply with business requirements.

    \item \textbf{5 points:} The response is fully relevant. It directly addresses the caller’s content, strictly follows business requirements, and all information provided is appropriate and consistent with the current dialogue context.
\end{itemize}

\textbf{Step 3.} Based on the above criteria and the full dialogue context, assign a reasonable score to the callee’s latest response only.  
Only output the final score in the required format without any additional explanation.
\end{tcolorbox}

\caption{A core prompt for response content relevance evaluation in intelligent call assistant scenarios. 
The rubric provides a clear definition of content relevance, explicit 1–5 point scoring criteria, and step-by-step evaluation instructions, enabling consistent and fine-grained assessment of the callee’s latest response.}
\label{fig:content-relevance-guidelines}
\end{figure*}

\end{document}